\PassOptionsToPackage{square,comma,numbers,sort&compress}{natbib}
\documentclass{article}

\usepackage[final]{neurips_2021}

\usepackage[utf8]{inputenc} 
\usepackage[T1]{fontenc}    
\usepackage{hyperref}       
\usepackage{url}            
\usepackage{booktabs}       
\usepackage{amsfonts}       
\usepackage{nicefrac}       
\usepackage{microtype}      
\usepackage{xcolor}         

\usepackage{authblk}

\usepackage{amsmath}

\usepackage{physics}
\usepackage{graphicx}
\usepackage{multirow}
\usepackage{tabularx}
\usepackage{float}
\hypersetup{
    colorlinks=true,
    linkcolor=black,
    citecolor=black,
    urlcolor=blue,
    }

\title{Explainable Semantic Space by Grounding Language to Vision with Cross-Modal Contrastive Learning}

\author[1,2]{Yizhen Zhang}
\author[1]{Minkyu Choi}
\author[1]{Kuan Han}
\author[1,3]{Zhongming Liu}
\affil[1] {
  Department of Electrical Engineering and Computer Science\\
  University of Michigan\\
  Ann Arbor, MI 48109}
\affil[2] {
 Department of Neurological Surgery\\
 University of California San Francisco\\
 San Francisco, CA 94143}
\affil[3] {
  Department of Biomedical Engineering\\
  University of Michigan\\
  Ann Arbor, MI 48109}
\affil[ ]{\ttfamily\{zhyz, cminkyu, kuanhan, zmliu\}@umich.edu}

\begin{document}

\maketitle

\begin{abstract}
In natural language processing, most models try to learn semantic representations merely from texts. The learned representations encode the “distributional semantics” but fail to connect to any knowledge about the physical world. In contrast, humans learn language by grounding concepts in perception and action and the brain encodes “grounded semantics” for cognition. Inspired by this notion and recent work in vision-language learning, we design a two-stream model for grounding language learning in vision. The model includes a VGG-based visual stream and a Bert-based language stream. The two streams merge into a joint representational space. Through cross-modal contrastive learning, the model first learns to align visual and language representations with the MS COCO dataset. The model further learns to retrieve visual objects with language queries through a cross-modal attention module and to infer the visual relations between the retrieved objects through a bilinear operator with the Visual Genome dataset. After training, the model’s language stream is a stand-alone language model capable of embedding concepts in a visually grounded semantic space. This semantic space manifests principal dimensions explainable with human intuition and neurobiological knowledge. Word embeddings in this semantic space are predictive of human-defined norms of semantic features and are segregated into perceptually distinctive clusters. Furthermore, the visually grounded language model also enables compositional language understanding based on visual knowledge and multimodal image search with queries based on images, texts, or their combinations.

\end{abstract}

\section{Introduction} \label{sec:introduction}

Humans take much longer time to name a colored word when the color and the word mismatch (e.g., “\emph{red}” shown in green) than when they match (e.g., “\emph{red}” shown in red) \citep{stroop1935studies}. This effect is an example of rich psychological evidence suggesting that humans learn language by grounding meanings to knowledge about the world \citep{pulvermuller2013neurons, gunther2020immediate}. In contrast, most models in natural language processing (NLP) \citep{mikolov2013distributed, sutskever2014sequence, devlin2019bert, brown2020language} encode “distributional semantics” \citep{emerson2020goals} learned from texts only. Put yourself as machines in a thought experiment for the “Chinese Room Argument” \citep{searle1980minds}. Imagine that you have to learn Chinese from scratch as your first language. All that you have is a Chinese-to-Chinese dictionary. You might be able to relate a word to other words based on textual distributions. It is, however, impossible to learn word meanings without any additional explanation in reference to the physical world \citep{harnad1990symbol}.

A language model may learn concepts from texts paired with sensory data, such as images. Joint vision-language learning has been explored for image captioning \citep{lin2014microsoft}, visual question answering \citep{hudson2019gqa}, and pre-training vision models with weak supervision \citep{jia2021scaling,radford2021learning}. In line with these studies, we train a language model and a vision model jointly to match images and texts. We further analyze the semantic space obtained with the visually grounded language model. In this space, semantic embeddings are found to be organized and clustered by visual attributes, predictive of human-defined norms of semantic features, useful for compositional language understanding and cross-modal image search. We expect this visually grounded language model to also be useful for understanding the computational basis of grounded cognition \citep{martin2016grapes, barsalou2008grounded}.

\begin{figure}[htp]
  \centering
  \includegraphics[width=\textwidth]{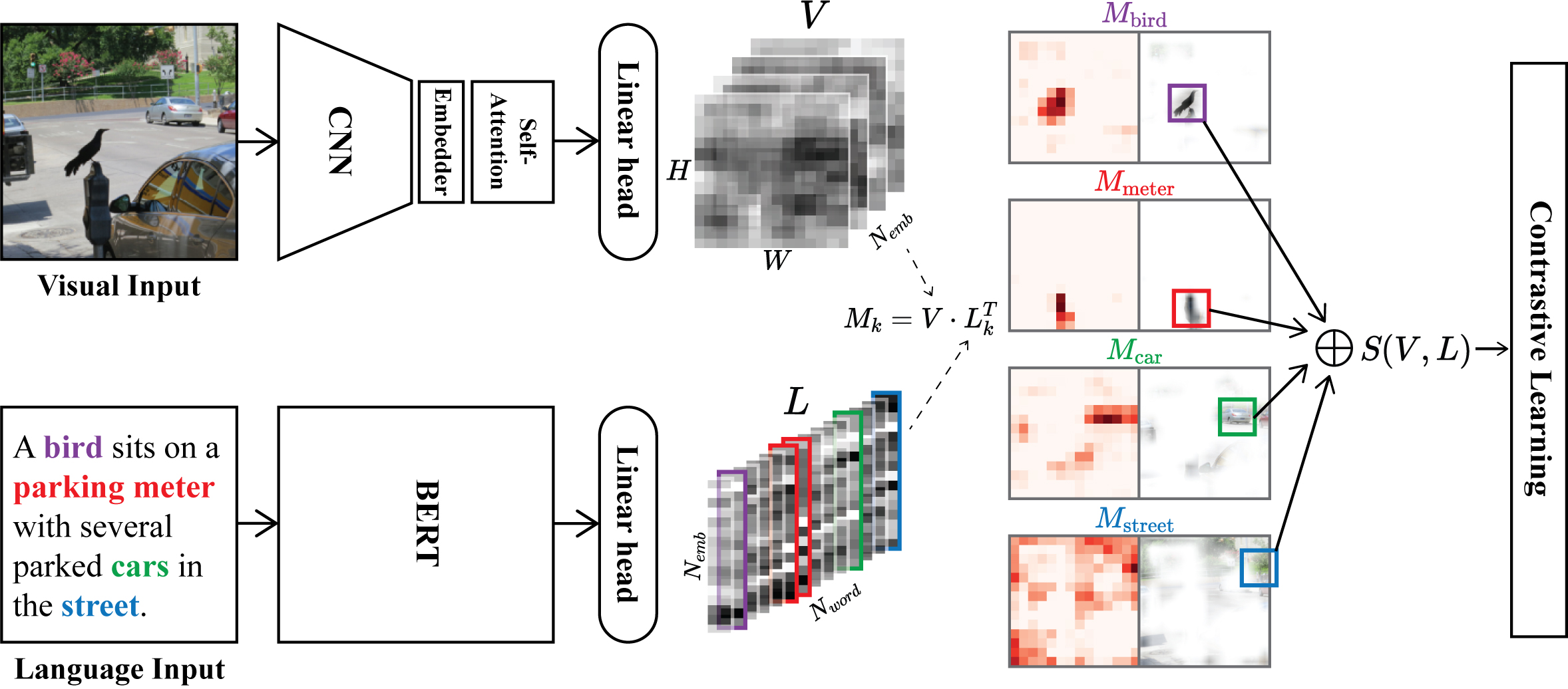}
  \caption{Visual grounding of natural language (see Section~\ref{sec:visual_grounding_natural_language}). The visual and language streams take an image and its caption as input, respectively. The inner-product between the visual feature maps and the contextual word embeddings forms the 3D match-map that highlights the matching between visual and language content. The similarity score calculated from the match-map (see Eq.~\ref{eq:matchmap}) is used to evaluate the cross-modal contrastive loss.}
  \label{fig:pretrain_structure}
\end{figure}

\section{Background and Related Work} \label{sec:background}

\subsection{Distributional vs. grounding hypothesis} \label{sec:language_learning_principle}

In the \textit{distributional hypothesis} \cite{harris1954distributional}, words that occur in similar contexts carry similar meanings. This hypothesis has motivated influential machine learning models to learn word embeddings from large text corpora \citep{mikolov2013distributed, pennington2014glove}. However, the learned word embeddings are not straightforward to interpret \citep{bisk2020experience, emerson2020goals}. Alternatively, the \textit{symbol grounding hypothesis} suggests that a word is connected to its meaning by relating to its referent in the physical world \citep{harnad1990symbol, glenberg2000symbol}. In line with this hypothesis, earlier studies demonstrate that visual features or contexts can enhance language learning \citep{silberer2012grounded, roller2013multimodal, bruni2014multimodal, hill2014multi, zablocki2018learning}. 

\subsection{Vision-language learning} \label{sec:visual_language_learning}

Grounding language in vision has been of increasing interest in computational linguistics and machine learning. A common strategy is to fuse words with related visual information in terms of perceptual norms \citep{silberer2012grounded}, bag-of-visual-word \citep{bruni2014multimodal,lazaridou2014wampimuk}, or learnable visual features \citep{lazaridou2015combining, chrupala2015learning, kiros2018illustrative, ailem2018probabilistic}. The models used for vision-language fusion evolve alongside those for NLP, such as Latent Dirichlet Allocation (LDA) \citep{bruni2014multimodal}, log-bilinear model \citep{gupta2019vico}, Skip-gram model \cite{lazaridou2015combining, zablocki2018learning}, and recurrent neural network \citep{chrupala2015learning}. Generally, visual grounding may refine the distribution and interpretability of language representations \citep{bruni2014multimodal, lazaridou2014wampimuk,chrupala2015learning,kiros2018illustrative,zablocki2018learning,ailem2018probabilistic,gupta2019vico,bordes2020incorporating} and facilitate cross-modal tasks \citep{chrupala2015learning,kiros2018illustrative,ailem2018probabilistic,gupta2019vico,bordes2020incorporating,baroni2016grounding}. More recent work has begun to use transformer \citep{vaswani2017attention, devlin2019bert} for vision-language learning, showing strong performance in cross-modal tasks \citep{tan2019lxmert, lu2019vilbert, su2019vl, chen2019uniter} .

Contrastive learning \citep{becker1992self, bromley1993signature} is increasingly applied to not only unimodal data \citep{chen2020simple} but also multimodal data \citep{harwath2018jointly, jia2021scaling}. It is able to learn better representations than alternative prediction or classification objectives \citep{tian2020contrastive}. However, cross-modal contrastive learning is still under-explored for higher-level tasks, e.g., visual question answering \citep{hudson2019gqa}, visual reasoning \citep{suhr2019corpus}, scene graph generation \citep{krishna2017visual}. Such tasks involve abstract reasoning about the relations between entities (e.g., visual objects). Prior work approaches relational inference with multi-layer perceptron \citep{lu2016visual, santoro2017simple} or graph neural networks \citep{chen2018iterative, li2019relation, li2019visual}. Arguably, a more compelling idea \citep{hakami2017pairdiff} is to model entities as vectors in a continuous space and to model their relations as arithmetic operators (linear \citep{mikolov2013linguistic, bordes2013translating, wang2014knowledge} or bilinear \citep{yang2014embedding, nickel2016holographic}) applied to the vector representations of those entities.

\subsection{Relation to prior work} \label{sec:relation_to_prior_work}

In this work, we first build a two-stream model to jointly learn visual and language representation from image-caption pairs, similar to recent work  \citep{jia2021scaling,radford2021learning}. We then finetune the learned model by adding a cross-modal attention layer \citep{lu2019vilbert, tan2019lxmert} and bilinear operators \citep{yang2014embedding} to represent the relations between visual objects. Both stages utilize cross-modal contrastive loss. Related to our work, Harwath et al. match visual objects to spoken words using triplet loss \cite{harwath2018jointly}. Early this year, Jia et al. \citep{jia2021scaling} and Radford and Kim et al. \citep{radford2021learning} use contrastive learning to pretrain a vision model using a massive image-text dataset and demonstrate largely improved zero-shot transfer learning performance on visual and cross-modal tasks. Different from their perspectives, we focus on assessing the language encoders and word representations. Specifically, we perform a systematic evaluation of the semantic space grounded in vision vs. the ungrounded semantic space learned from texts only. This evaluation is possible since after training, the language and visual streams in our model are fully separable as stand-alone systems, unlike some vision-language models that require both visual and textual input to be usable \citep{lu2019vilbert, tan2019lxmert}. Our goal is to assess how visual grounding affects the distribution of textual representations by analyzing the distribution of word embeddings in the grounded semantic space, in line with related works \citep{bruni2014multimodal, lazaridou2014wampimuk,chrupala2015learning,kiros2018illustrative,zablocki2018learning,ailem2018probabilistic,gupta2019vico,bordes2020incorporating,ilharco2020probing} . 

\section{Approach}  \label{sec:approach}

\subsection{Visual grounding of natural language}  \label{sec:visual_grounding_natural_language}
To build a computational model for learning visually grounded language representations, we develop a model (Fig.~\ref{fig:pretrain_structure}) that combines a stand-alone visual stream and a stand-alone language stream. The visual stream is based on VGG16 \citep{simonyan2014very} with an additional linear transformation as an embedder to match the feature dimension of the language stream and an additional multi-head self-attention layer \citep{carion2020end} to enforce global information aggregation and learn long-range dependency. The language stream is based on Bert \citep{devlin2019bert}. Using separate linear transformation heads \citep{chen2020simple}, the output from both the visual stream and the language stream are projected to a common representational space. In this common space, the inner-product between the visual representation $\vb*{V}$ at every location and the language representation $\vb*{L}$ of every word gives rise to a 3D match-map, where each element indicates how a word in the text matches each location in the image (See illustration in Fig.~\ref{fig:pretrain_structure}). The sum of the maximal match is the similarity score  $S(\vb*{V}, \vb*{L})$ between a pair of image and text. See Eq.~\ref{eq:matchmap}, where $i,j$ indicate the location in the 2D image feature map $\vb*{V}$ and $k$ indicates the $k$-th word in $\vb*{L}$.
\begin{equation}
    \small
    \label{eq:matchmap}
    \vb*{M}_{i,j,k} = \vb*{V}_{i,j} \cdot \vb*{L}_k^T, \quad
    S(\vb*{V}, \vb*{L}) = \sum _{k=1}^K \max_{i,j} \vb*{M}_{i,j,k}
\end{equation}
Extending the unimodal normalized temperature-scaled cross-entropy (NT-Xent) loss \citep{oord2018representation, chen2020simple, jia2021scaling}, we define the cross-modal contrastive loss using the anchor sample from one modality and the positive sample and negative samples from the other modality \citep{jia2021scaling,radford2021learning}. As such, we define and sum two loss functions with the anchor sample from either images or texts and positive/negative samples from either texts or images, respectively. 
\begin{equation}
    \small
    \text{Loss}_l = -\frac{1}{B} \sum_{i=1}^B \log \frac{\exp(S(\vb*{V}_i, \vb*{L}_i)/\tau)}{\sum_{j=1}^B \exp(S(\vb*{V}_i, \vb*{L}_j)/\tau)}, \quad
    \text{Loss}_v = -\frac{1}{B} \sum_{i=1}^B \log \frac{\exp(S(\vb*{V}_i, \vb*{L}_i)/\tau)}{\sum_{j=1}^B \exp(S(\vb*{V}_j, \vb*{L}_i)/\tau)}
    \label{eq:pretrain_loss}
\end{equation}

For $\text{Loss}_l$ in Eq.~\ref{eq:pretrain_loss}, the anchor sample $\vb*{V}_i$ is an input image and the positive sample $\vb*{L}_i$ is the corresponding image caption, whereas the negative samples $\vb*{L}_j$ are unmatched textual descriptions included in the same batch ($B$ is the batch size). Similarly, $\text{Loss}_v$ in Eq.~\ref{eq:pretrain_loss} is defined to contrast the positive and negative image samples against an anchor textual sample.

\subsection{Visual grounding of object relations}   \label{sec:visual_grounding_object_relation}

\begin{figure}[htp]
  \centering
  \includegraphics[width=0.9\textwidth]{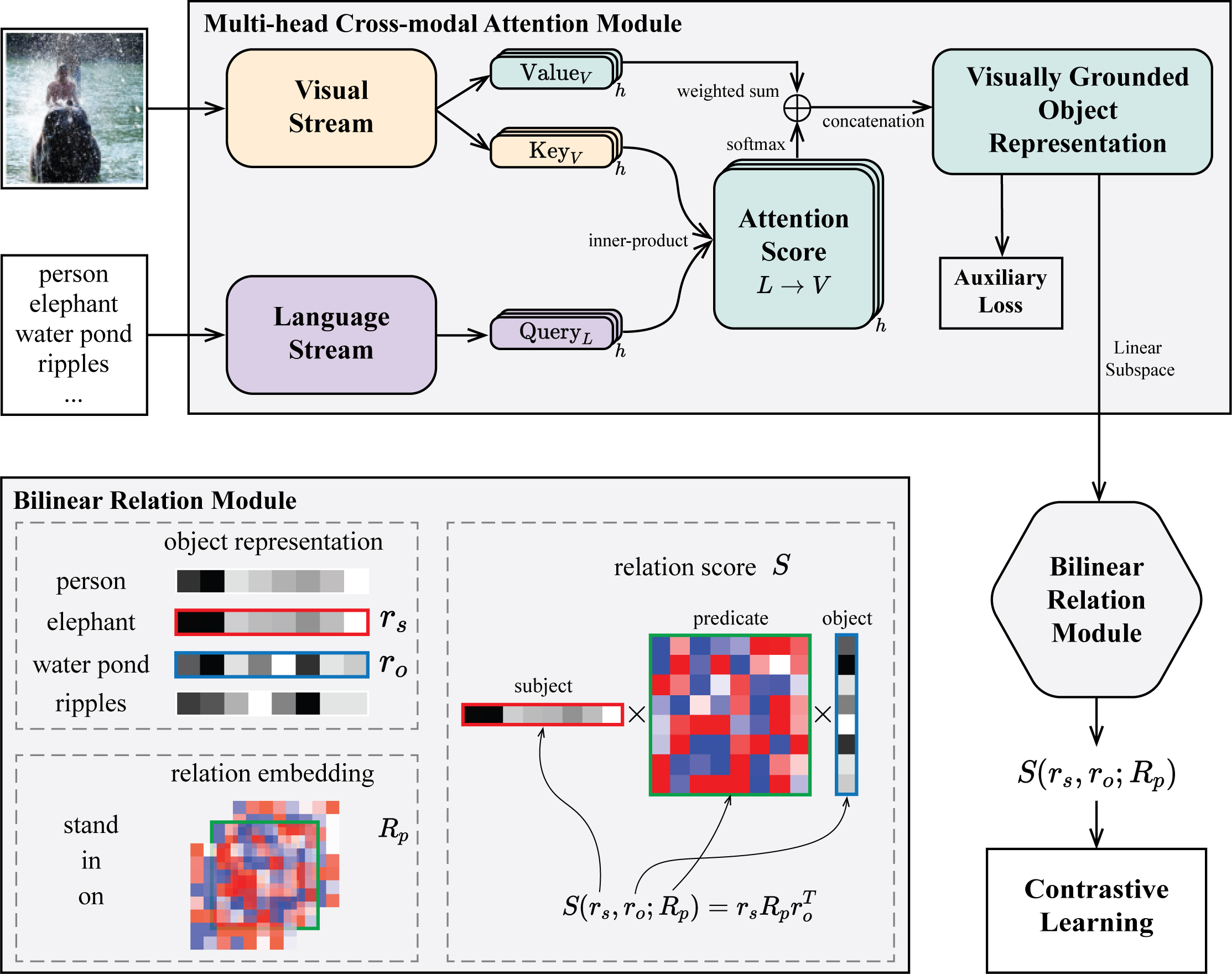}
  \caption{Visual grounding of object relation. The language stream uses an object description as input (e.g., \texttt{\small{large black elephant}}; we only show the object name "elephant" in this illustration for simplicity). The multi-head cross-attention module outputs a set of visually grounded object representations (See detailed methods in Appendix A.3). The bilinear relation module (bottom left) further generates a relation score given representations of a (\textbf{s}ubject,\textbf{p}redicate,\textbf{o}bject) triplet (e.g. \texttt{\small{(elephant,in,water pond)}}) for contrastive learning.}
  \label{fig:finetune_structure}
\end{figure}

We further finetune the model for visual relation prediction, as illustrated in Fig.~\ref{fig:finetune_structure}. In this stage, we remove the linear transformation heads in Fig.~\ref{fig:pretrain_structure} and add a multi-head cross-modal attention module \citep{lu2019vilbert, tan2019lxmert}. The attention module uses a query based on the embedding of an object word from the language stream (Query$_L$) and uses keys (Key$_V$) and values (Value$_V$) from every location in the feature map output from the visual stream. The attention score is calculated as the inner-product of Query$_L$ and Key$_V$ followed by softmax. The attention-weighted sum of Value$_V$ is concatenated across 8 attention heads to generate a visually grounded object representation.

Applied to the grounded object representation is a bilinear relation module for predicting the visual relation between two objects (linguistically a subject and an object). For both the subject and the object, their grounded representations are linearly transformed to a subspace $D = \mathbb{R}^d$ (default $d=32$), denoted as $r_s$ and $r_o$, respectively. A predicate $p$ is represented as a learnable bilinear operator $F_p: D \times D\to \mathbb{R}$, which represents the relation embedding $R_p$. Applying this bilinear operator to the subject vs. object representations measures their relation score $S$ specific to the given predicate \citep{yang2014embedding} expressed as (Eq.~\ref{eq:bilinear_operator}). 
\begin{equation}
    \small
    S(r_s, r_o; R_p) = F_p(r_s, r_o) = r_s R_p r_o^T
    \label{eq:bilinear_operator}
\end{equation}
For visual relation prediction, we also use contrastive learning with two loss functions by taking either relation embedding or subject/object representations as positive/negative samples.
\begin{equation}
    \small
    \text{Loss}_{\text{rel}} = -\frac{1}{|\mathcal{B}|} \sum_{(r_s, r_o;R_p) \in \mathcal{B}} \log \frac{\exp(S(r_s, r_o; R_p)/\tau)}{\sum_{k \in \mathcal{K}_{\text{rel}}} \exp(S(r_s, r_o; R_p^k)/\tau)}
    \label{eq:finetune_loss_rel}
\end{equation}

\begin{equation}
    \small
    \text{Loss}_{\text{obj}} = -\frac{1}{|\mathcal{B}|} \sum_{(r_s, r_o;R_p) \in \mathcal{B}} \log \frac{\exp(S(r_s, r_o; R_p)/\tau)}{\sum_{k \in \mathcal{K}_{\text{obj}}} \exp(S(r_s^k, r_o^k; R_p)/\tau)}
    \label{eq:finetune_loss_obj}
\end{equation}

In $\text{Loss}_{\text{rel}}$, $\mathcal{K}_{\text{rel}}$ is the set that contains all relations available. The anchor sample is a pair of subject and object in an image. The positive sample is the embedding of the ground truth relation. The negative samples are the embeddings of all other relations. In $\text{Loss}_{\text{obj}}$, the anchor sample is a given relation. The positive sample is a subject-object pair that holds this relation. The negative samples are other subject-object pairs in a different relation. For both loss functions, the positive and negative samples are drawn from the same batch $\mathcal{B}$. 

In addition, we also add a classification head (two fully connected layers with ReLU in between) and apply it to the grounded object representation. We use object classification as an auxiliary objective (with a cross-entropy loss) to constrain the grounded object representation to be separable across objects for classification. 

\subsection{Training and Testing}  \label{sec:training_testing}

We train the model in three stages to progressively refine the model with increasingly demanding tasks. In the first stage, we pretrain the visual and language streams separately as image and text encoders. The language stream is the pretrained Bert\footnote{\texttt{bert-base-uncased}: \url{ https://huggingface.co/transformers/pretrained_models.html}}  used as the baseline model for subsequent experiments. The visual stream is pretrained for object classification with ImageNet \citep{russakovsky2015imagenet}.  Relative to the baseline CNN, the inclusion of self-attention improves the top-1 classification accuracy from $71.6 \%$ to $74.3 \% $ on the ImageNet validation dataset. The attention module also renders the classification more robust when the input image is partially occluded (See details in Appendix A.1). 

In the second stage, we refine the pretrained language and visual streams by matching texts to images, as illustrated in Fig.~\ref{fig:pretrain_structure} on the MS COCO dataset \citep{lin2014microsoft}. While freezing other layers, we refine the self-attention layer in the visual stream and the top $k$ layers in Bert (by default $k=8$). Training with contrastive learning is based on the MS COCO dataset. As five captions are available for each image, we randomly sample one caption per image in each iteration. Earlier grounding (larger $k$) tends to support better image-text retrieval performance (see details in Appendix A.2).

In the third stage, we further finetune the model for visual relation prediction as illustrated in Fig.~\ref{fig:finetune_structure}. We refine the visual self-attention layer and the higher $l$ layers in Bert (by default $l=2$) based on the Visual Genome dataset \citep{krishna2017visual} after cleaning the dataset to include $114$ relations and $55$ object classes in order to alleviate imbalanced data across different classes or relations (Appendix A.3). The training does not use any image annotation (e.g., bounding box) , which otherwise requires other models (e.g., object detection). Instead, we use cross-attention to retrieve visual objects from raw images given textual queries and learn the representations of the retrieved objects and their relations altogether. After training, the model predicts the object class with $97.71\%$ top-1 accuracy and predicts the relation label with $64.26\%$ top-1 accuracy (See details and examples in Appendix A.3).

\section{Experiments}  \label{sec:experiment}

\subsection{Principal components of grounded semantic representations}  \label{sec:pca}

To evaluate the visually grounded semantic space, we use the language stream as a stand-alone model to extract the output representations of commonly used English words in the SemCat dataset ($9,197$ words; $100$ word categories) \citep{csenel2018semantic}. Details about how word representations are extracted from the language stream are explained in Appendix B. We apply the principal component analysis to the representations of all the words studied here and examine the top components as the principal dimensions of the grounded semantic space. 

\begin{figure}[htp]
  \centering
  \includegraphics[width=\textwidth]{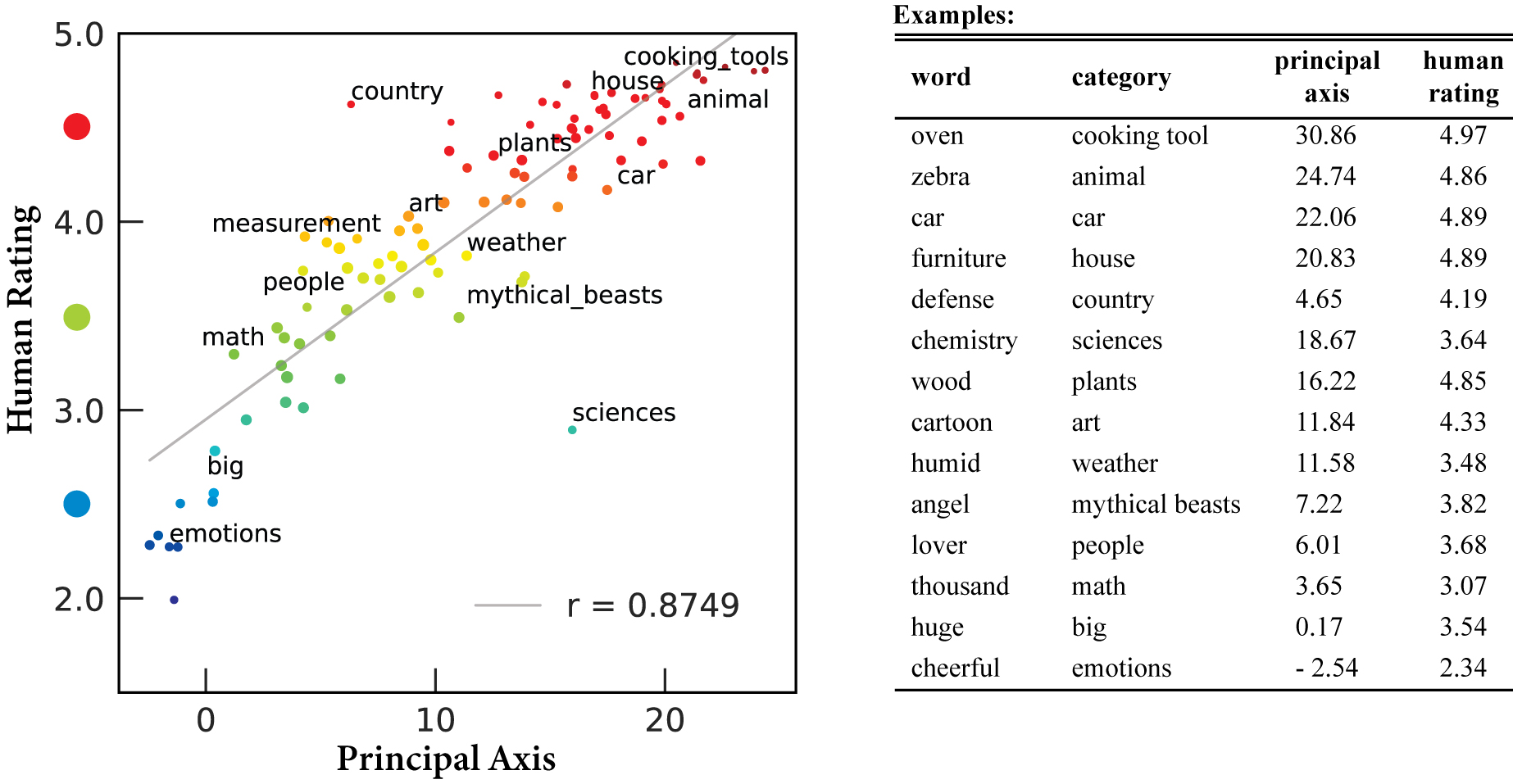}
  \caption{The first principal component in the grounded semantic space captures the concrete-abstract axis of semantics. Left: Each dot represents a word category with the color indicative of the averaged human-rated concreteness (the y axis) and the size proportional to the standard deviation. The x axis indicates the value projected onto the first principal axis. Right: Example words in labeled categories.}
  \label{fig:concreteness}
\end{figure}

\begin{table}[H]
  \small
  \caption{Correlation between the $1^{\text{st}}$ principal axis and human-rated word concreteness}
  \label{table:concreteness}
  \centering
  \begin{tabular}{m{2cm} m{1.5cm} m{1.5cm} m{1.5cm}}
    \toprule
     & \multicolumn{3}{c}{Correlation (Pearson's r)}  \\
    \cmidrule(r){2-4}
    Group           & Bert      & Grounded      & \small{Relational Grounded} \\
    \midrule
    word-level      & 0.1040    & 0.6615        & \textbf{0.6948} \\
    category-level  & 0.3538    & \textbf{0.8749}        & 0.8001 \\
    \bottomrule
  \end{tabular}
\end{table}

Interestingly, the first principal dimension is readily interpretable as an abstract-to-concrete axis (Fig.~\ref{fig:concreteness}). For example, words with the highest values in this axis are \emph{ostrich, seagull, albatross, blender, pelican, broccoli, parakeet, lettuce, sailboat, vegetables}, whereas words with the lowest values are \emph{displeasure, liking, to, outgoing, present, experienced, profitable, faithful, meaningful, multitude}. The representations of words along this axis is significantly correlated with human rating of their concreteness (ranging from 1 to 5) from prior study \citep{brysbaert2014concreteness} (Fig.~\ref{fig:concreteness}). The Pearson correlation coefficient reaches $0.8749$ or $0.6615$ across word categories or individual words, respectively; after grounding with object relations: $r=0.8001$ for categories, $r=0.6948$ for words. In contrast, the principal axis of the ungrounded semantic space learned from the baseline Bert model is not straightforward to interpret and shows a weak correlation with human ratings of concreteness (Table.~\ref{table:concreteness}). Other principal components are also intuitively interpretable. For example, PC $2$ captures the human vs. non-human axis, PC $3$ captures the scene vs. object axis, PC $4$ captures the natural vs. artificial axis, PC $5$ captures the indoor vs. outdoor axis, PC $6$ highlights words related to food. See results about other principal components in Appendix B.1.

\subsection{Relation to human-defined norms of semantic features}   \label{sec:semantic_norm}

\begin{figure}[htp]
  \centering
  \includegraphics[width=0.9\textwidth]{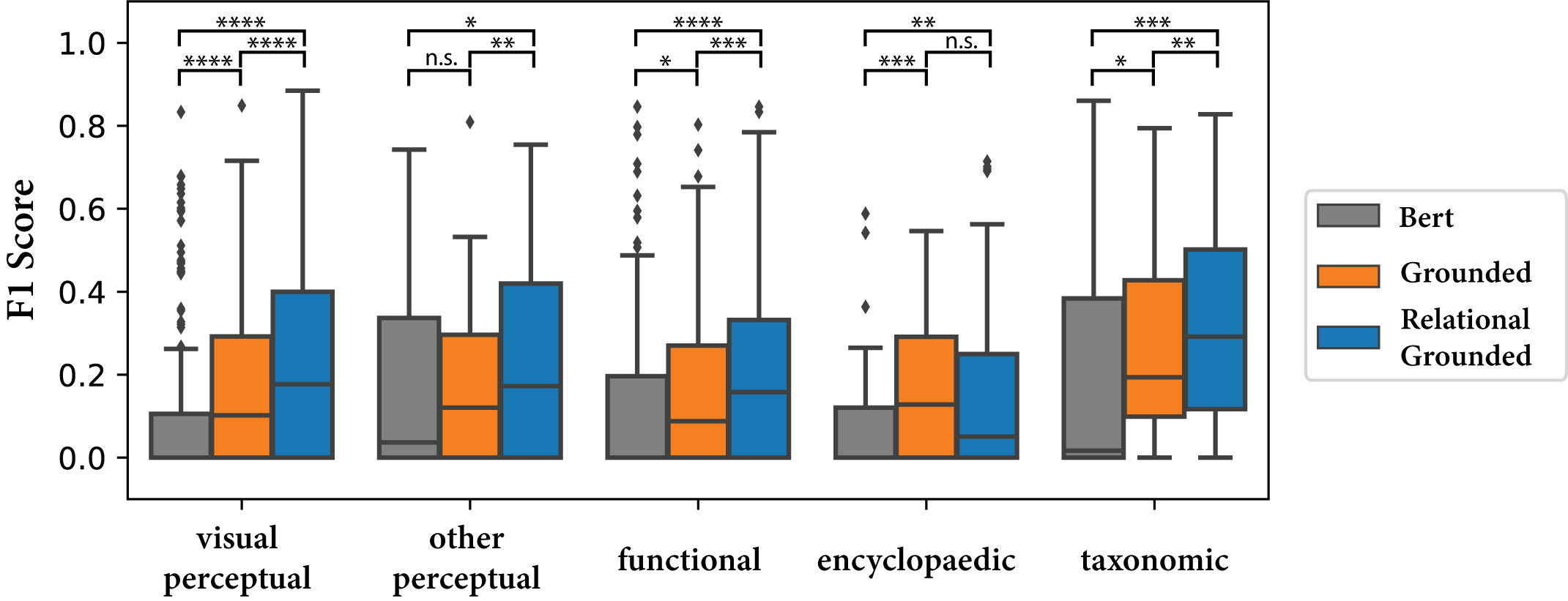}
  \caption{The F1 score of predicting semantic feature norms from word representations before and after visual grounding. Each box shows the lower ($25\%$) percentile, the higher ($75\%$) percentile, and the median of F1 scores within a feature type. Whisker$=1.5$. Significant level: n.s.: not significant; *, $p<0.05$; **, $p<0.01$; ***, $p<0.001$; ****, $p<0.0001$.}
  \label{fig:semantic_norm}
\end{figure}
We further ask whether the visually grounded word embeddings are amenable to binary semantic features defined by humans \citep{li2017distributional, zablocki2018learning}. We use the concept property norm dataset from the Centre for Speech, Language and the Brain (CSLB) \citep{devereux2014centre}. The dataset includes binary semantic features (e.g., \texttt{has\_wheels}) labeled for $638$ concepts collected from $123$ human participants. We keep $390$ features that each contains at least 5 samples. We hypothesize that the grounded word embeddings can be readout with a linear and sparse projection to readily support binary classification attainable by humans. To test this hypothesis, we train a logistic regression model with L1 regularization to predict each binary semantic feature from the grounded word embeddings and also repeat this for ungrounded semantics for comparison. See Appendix B.2 for details about this dataset and our evaluation method. Results suggest that the grounded word embeddings are significantly more predictive of visually relevant binary features than ungrounded counterparts obtained by Bert (Wilcoxon Signed Rank Test; $p<0.0001$) (Fig.~\ref{fig:semantic_norm}). This difference is less pronounced but still significant for other features related to other perceptual (e.g., \texttt{has\_flavors}), functional (e.g., \texttt{does\_cut}), encyclopaedic (e.g., \texttt{is\_dangerous}), and taxonomic features  (e.g., \texttt{is\_clothing}), especially after visual grounding of object relations. 

\subsection{Clustering of word representations} \label{sec:categorization}
\begin{figure}[htp]
  \centering
  \includegraphics[width=\textwidth]{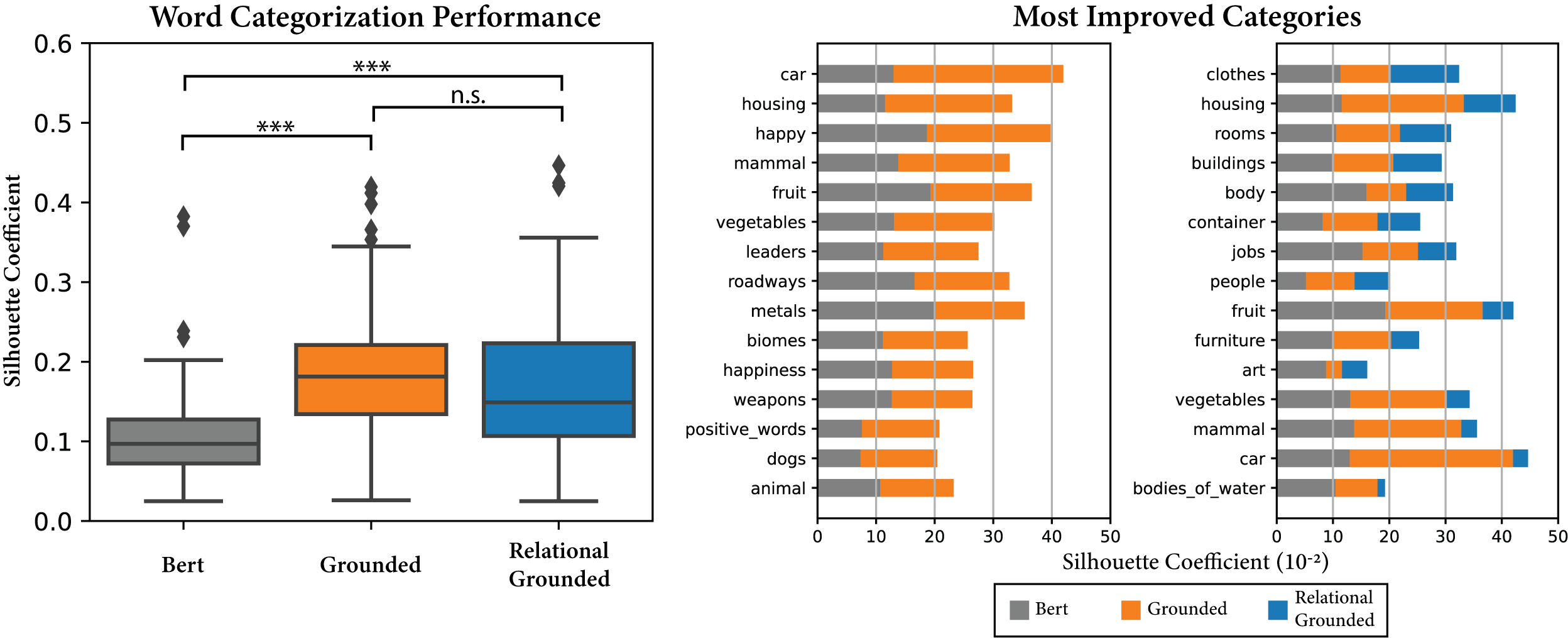}
  \caption{Left: A boxplot showing Silhouette coefficients on word representations before and after visual grounding. Each box shows the lower ($25\%$) percentile, the higher ($75\%$) percentile, and the median of the Silhouette coefficients. Whisker$=1.5$.  Right: Top-$15$ word categories that are better clustered after visual grounding of natural language and object relations.}
  \label{fig:clustering}
\end{figure}

After visual grounding, the semantic representations tend to group themselves based on perceptual similarity. We use the SemCat dataset ($9,197$ English words from $N=100$ categories) \citep{csenel2018semantic} and calculate the Silhouette coefficient (between $-1$ and 1) to measure the degree to which these words are clustered by categories. The distance between word embeddings is measured as the cosine distance (See details in Appendix B.3). The Silhouette coefficients across 100 categories are significantly higher for the visually grounded semantics than ungrounded ones (Wilcoxon Signed Rank Test; $p<0.0001$) (Fig.~\ref{fig:clustering} left). The greatest gain in clustering are noticeable for categories that include concrete concepts (e.g. \texttt{car}, \texttt{housing}, \texttt{mammal}) with defining visual attributes (Fig.~\ref{fig:clustering} right). For some abstract categories related to human emotion (e.g., \texttt{happy}), the grounded representations are also better clustered than the ungrounded ones. 

\subsection{Visually informed compositional reasoning}   \label{sec:compositional}

\begin{figure}[htp]
  \centering
  \includegraphics[width=0.9\textwidth]{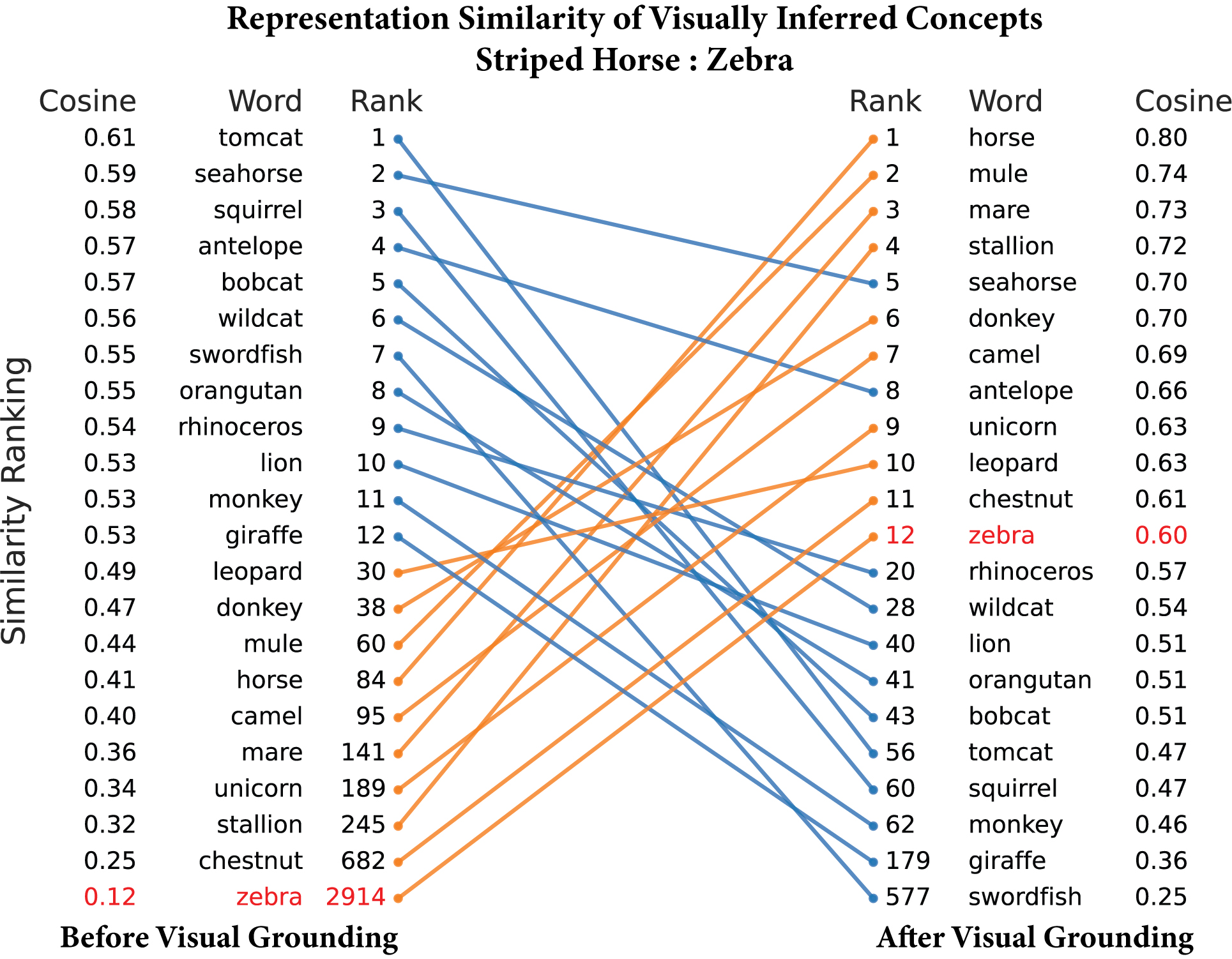}
  \caption{Compositional reasoning (\textbf{striped horse}). The left part shows the cosine similarity and its ranking between each of the listed words and the query phrase (striped horse) before visual grounding. The right part shows the corresponding results after visual grounding of natural language. Orange lines indicate words with increased ranking after visual grounding and blue lines for the decreased cases. We highlight in red the target word "zebra" for this specific example, which shows a significant increase in cosine similarity (from $0.12$ to $0.60$) and ranking (from $2914$ to $12$ out of $6238$ unique words). Besides, the top words similar to "striped horse" are all horse-like animals after visual grounding, but this is not the case for the ungrounded Bert model.}
  \label{fig:visual_inference_striped_horse}
\end{figure}

A drawback of distributional semantics is the inability to make visually informed compositional reasoning. We know that "zebra is a horse with black and white stripes", because we have seen how zebra looks like, whereas an ungrounded language model is never or rarely exposed to such information \citep{harnad1990symbol}. We test whether the visually grounded semantics can perform compositional reasoning based on visual knowledge, without being explicit trained to do so. We choose some words (Table~\ref{table:visual_inference}), for which the meaning can be intuitively inferred from the combination of other words.

\begin{table}[htp]
  \small
  \caption{Examples of visually informed conceptual composition. Each row shows the cosine similarity and its ranking in the vocabulary (unique words in the Semcat dataset) between the query phrase and the target word. Except (\texttt{hot weather}, \texttt{summer}), all others are concepts supported by composition of \emph{visual} knowledge in the query phrase. For each case, the highest similarity are rank are in bold.}
  \label{table:visual_inference}
  \centering
  \begin{tabular}{m{3cm} m{2cm} m{0.8cm} m{0.8cm} m{0.8cm} m{0.8cm} m{0.8cm} m{0.8cm}}
    \toprule
    \multirow{2}{*}{\textbf{Query Phrase}} & \multirow{2}{*}{\textbf{Target Word}} & \multicolumn{6}{c}{\textbf{Similarity (cosine  |  rank)}} \\
    \cmidrule(r){3-8}
    && \multicolumn{2}{c}{Bert} & \multicolumn{2}{c}{Grounded} & \multicolumn{2}{c}{Relational} \\
    \hline
    striped horse           & zebra         & 0.12	& 2914  & 0.60	& 12    & \textbf{0.63}	& \textbf{8}   \\
    black and white bear    & panda         & 0.13	& 2478  & 0.69	& \textbf{2}     & \textbf{0.81}	& \textbf{2}   \\
    flying car              & plane         & 0.36	& 167   & \textbf{0.66}	& \textbf{4}     & 0.61	& 11  \\
    round container         & bowl          & 0.25	& 489   & 0.56	& 8     & \textbf{0.67}	& \textbf{2}   \\
    red fruit               & strawberry    & 0.39	& 239   & 0.75	& \textbf{3}     & \textbf{0.85}	& \textbf{3}   \\
    young dog               & puppy         & 0.40	& 94    & 0.92	& \textbf{2}     & \textbf{0.93}	& \textbf{2}   \\
    iced mountain           & glacier       & 0.44	& 20    & \textbf{0.86}	& \textbf{1}     & 0.73	& 5   \\
    clear sky               & sunny         & 0.27	& 631   & 0.31	& 184   & \textbf{0.34}	& \textbf{61}  \\
    \hline
    hot weather             & summer        & 0.27	& 903   & 0.52	& 14    & \textbf{0.53}	& \textbf{6}   \\
    \bottomrule
  \end{tabular}
\end{table}

For example, we use a phrase "striped horse" as a compositional query to search for the matched words ranked in terms of cosine similarity. In Fig.~\ref{fig:visual_inference_striped_horse}, the left part shows the cosine similarity and ranking between each of the listed words and the query phrase "striped horse" before visual grounding. The right part shows the corresponding results after visual grounding of natural language. With the grounded semantic representation, the phrase \texttt{striped horse} is highly similar to the word \texttt{zebra} (cosine similarity: $0.60$), which is ranked as the $12$-th in the vocabulary. After further grounding the language model with visual object relations,  the target word \texttt{zebra} has an even higher cosine similarity of $0.63$ ranked the $8$-th in the vocabulary (Table~\ref{table:visual_inference}). Other top-ranked words all refer to horse-like animals (i.e., \emph{horse, mule, mare, stallion, donkey, camel, antelope}). This is in sharp contrast to the ungrounded semantic space, in which it is impossible to relate \texttt{striped horse} to \texttt{zebra} based on the similarity of their representations (cosine similarity: 0.12; rank: 2,914). The ungrounded Bert model highlights the top-$3$ similar words as \emph{tomcat, seahorse, squirrel}, which are animals sharing fewer visual features with horse-like animals. See other examples in (Table~\ref{table:visual_inference} and Appendix B.4).

\subsection{Multimodal image search}   \label{sec:multimodal_search}

\begin{figure}[htp]
  \centering
  \includegraphics[width=\textwidth]{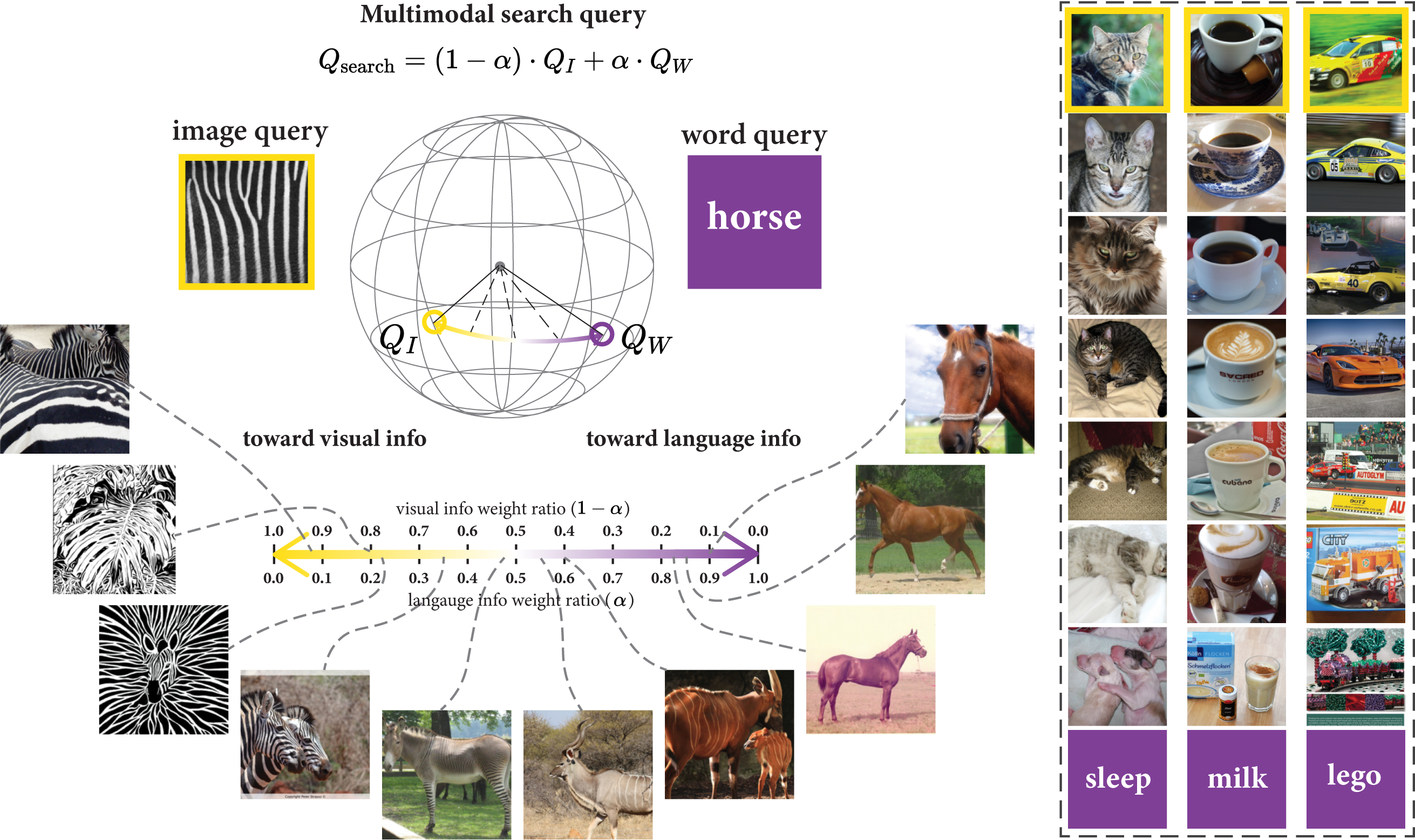}
  \caption{Left: Illustration of multimodal image search with a "zebra" example. The image query $Q_I$ is the L2-normalized vector representation of a zebra's skin pattern. The word query $Q_W$ is the L2-normalized vector representation of word \emph{horse}. As the weight ratio $\alpha$ in multimodal query $Q_{\text{search}}$ increases from $0$ to $1$, the search results show progressive changes from stripped patterns, to a real zebra, and to a horse image. Right: Example multimodal image search results using other image-word pairs as the combined query. The query images (highlighted with a yellow boundary) are in the top row and the corresponding words (shown in a purple background) are in the bottom row. The 1st to the 3rd columns correspond to combined queries, with a "cat" image and a \emph{sleep} word, a "coffee" image and a \emph{milk} word, a "car" image and a \emph{lego} word. The results of multimodal image search are shown in the 2nd to 6th rows, corresponding to increasing $\alpha$ from $0$ to $1$.}
  \label{fig:cross_moda_search}
\end{figure}

In our model, the cross-attention module forms a joint representational space to combine both visual and textual input. We explore whether this joint space can be used to support cross-modal tasks, e.g., image search based on image, text, or their combinations \citep{jia2021scaling}. For this task, we add two additional heads ($F_V$ and $F_L$). Each head includes two linear layers with ReLU in between followed by average pooling (See details in Appendix B.5). It is applied to either visual or textual representations in the joint space and results in a single vector representation for an image or a text (Eq.~\ref{eq:cross_modal_search_head}, $d=768$). While freezing our model described in Section~\ref{sec:visual_grounding_object_relation}, we train the two additional heads with contrastive loss to match the average-pooled representations of paired images and texts in terms of their cosine similarity using the MS COCO dataset. To use the model for image search, we apply weighted sum to the normalized representations of a query image and a query text (with weights: $1-\alpha$  and $\alpha$, where $0 \leq \alpha \leq 1$). We use this multimodal query (Eq.~\ref{eq:cross_modal_search_query}) to search a held-out database\footnote{$41,600$ images from the validation dataset of \href{https://storage.googleapis.com/openimages/web/index.html}{Open Images Dataset $V6$}.} for the matched images ranked in terms of cosine similarity.

\begin{equation}
    \small
    \label{eq:cross_modal_search_head}
    \vb*{Q}_I = F_V(\text{Key}_V) \in \mathbb{R}^d,  \quad
    \vb*{Q}_W = F_L(\text{Query}_L) \in \mathbb{R}^d.
\end{equation}

\begin{equation}
    \small
    \label{eq:cross_modal_search_query}
    \vb*{Q}_{\text{search}} = (1-\alpha) \vb*{Q}_I + \alpha \vb*{Q}_W.
\end{equation}

As $\alpha$ controls the weighting between the textual and visual queries, we test how the image search returns different results as $\alpha$ increases from $0$ (image only) to $1$ (text only). For example, when we combine a word (\textit{horse}) and an image (a stripped pattern) into a query, the search finds images similar to the zebra's skin pattern when $\alpha$ is close to $0$, or finds images of typical horses when $\alpha$ is close to $1$, but not necessarily a zebra for either case until when $\alpha$ is somewhere close to $0.5$ (Fig.~\ref{fig:cross_moda_search}, left). This observation is generalizable to other examples. See similarly graded changes in (Fig.~\ref{fig:cross_moda_search}, right).

\section{Discussion}   \label{sec:discussion}

In summary, we apply visual grounding to not only words but also relations between words through cross-modal contrastive learning. The results suggest that grounding language learning in vision renders semantic representations more interpretable by human intuition. The grounded semantic space has its principal dimension encode the concrete-to-abstract variation consistent with human ratings and neurobiological knowledge. The grounded semantic representations are better clustered by finer categories and capable of compositional reasoning (e.g., \texttt{zebra = striped horse}). In addition, our work also shows compelling evidence that both text and image-informed semantics are represented in a common, continuous, and grounded semantic space. Although this notion has been hypothesized in neuroscience and linguistics, it has been rarely implemented and demonstrated with computational models. Uniquely, we demonstrate that a continuously varying combination of a text and an image into a multimodal query can be used to search images, showing results that make intuitive sense.

Several limitations of our work are noteworthy. The datasets used to train our model are orders of magnitude smaller than those used in recent studies \citep{jia2021scaling,radford2021learning}. Scaling up the model training with increasingly larger datasets is expected to greatly improve the model’s performance for cross-modal tasks, while generally preserving the interpretability of the grounded language representations as described herein. Some of our experiments and results are preliminary and primarily for illustrative purposes and await more comprehensive and quantitative evaluation in future studies, especially with more downstream vision-language tasks. Whereas our evaluation focuses on the language model, grounding language to vision may also have refined the visual stream, awaiting further evaluation against visual tasks, as demonstrated in \citep{jia2021scaling,radford2021learning}.

The visually grounded language model may be usable as a computational model for studying the grounded cognition - a theory in cognitive science \citep{barsalou2008grounded, martin2016grapes}. Ungrounded linguistic models are explanatory about semantic processing in the brain's language  network \citep{mitchell2008predicting, huth2016natural, bulat2017speaking, pereira2018toward, zhang2020connecting}. Combining the grounded language model with human behavioral and neural data may elucidate how the language network interacts with distributed sensory and motor areas for semantic processing \citep{heinrich2020crossmodal}.

It is natural to extend this study by incorporating other sensory input \citep{kiela2015grounding, kiela2015multi} and further ground language learning in action \citep{yu2018interactive, chai2018language, paul2016efficient, lynch2020grounding} and emotion \citep{rotaru2020constructing}. This study also leaves an open question as to whether grounding should occur at an early or late stage of natural language processing, which awaits further exploration and evaluation. For comprehensive modeling of language grounding, it is desirable to expose an agent to a naturalistic and multi-sensory environment and to engage interactive actions to allow the agent to learn knowledge in the physical world, like how humans learn language.

\acksection   \label{sec:acksection}
Funding in direct support of this work: NSF IIS 2112773.

\bibliographystyle{unsrtnat}
{
\small
\bibliography{ms.bib}
}

\end{document}


\maketitle

\appendix

\section{Training and Testing}

\subsection{Visual stream pretraining}

\subsubsection{Training for ImageNet classification}

We pretrain the visual stream on ImageNet \citep{russakovsky2015imagenet} for object classification to evaluate whether adding a self-attention layer to VGG16 can help learn a better image representation. The linear embedder transforms the image feature from $512$ channels to $768$ channels to match the output feature dimension in the Bert model. This linear embedder also prepares the image and language representations for them to be merged in the second stage (see section~\ref{sec:training_second_stage}). In the visual stream, the single-layer self-attention has $12$ heads and follows the structure as described in the original transformer paper \citep{vaswani2017attention}. We add a 2D positional encoding \texttt{PE}$(x,y)$ to the input feature map before passing it to the self-attention layer. The positional encoding is learnable and includes $768 \times 14 \times 14$ parameters. We use the same hyper-parameter setting for training VGG16 and attention-enhanced visual stream (batch size$=200$, optimizer=\texttt{SGD}, learning rate$=0.01$, momentum$=0.9$, weight decay$=1\mathrm{e}{-4}$; learning rate decay by half for every $20$ epochs). The training is done with $4$ Nvidia GeForce GTX Titan XP Graphic Card ($12$ GB memory per card).

The  performance on ImageNet classification (see Table~\ref{table:ImageNet_pretrain_preformance}) is compared between VGG16 and its variation with attention enhancement. This result suggests that adding one additional self-attention layer on the top of VGG16 improves the classification performance on ImageNet. 

\begin{table}[htp]
  \caption{Object classification accuracy on ImageNet validation dataset.}
  \label{table:ImageNet_pretrain_preformance}
  \centering
  \begin{tabular}{m{3cm} m{1.5cm} m{1.5cm} m{1.5cm}}
    \toprule
     & \multicolumn{3}{c}{Object classification accuracy (\%)}  \\
    \cmidrule(r){2-4}
    Model               & Top-1     & Top-5     & Top-10 \\
    \midrule
    VGG16               & 71.6    & 90.4    & 94.0 \\
    VGG16$+$attention   & \textbf{74.3}    & \textbf{91.8}   & \textbf{95.1} \\
    \bottomrule
  \end{tabular}
\end{table}

\newpage
\subsubsection{Occlusion experiments}

\begin{figure}[H]
  \centering
  \includegraphics[width=\textwidth]{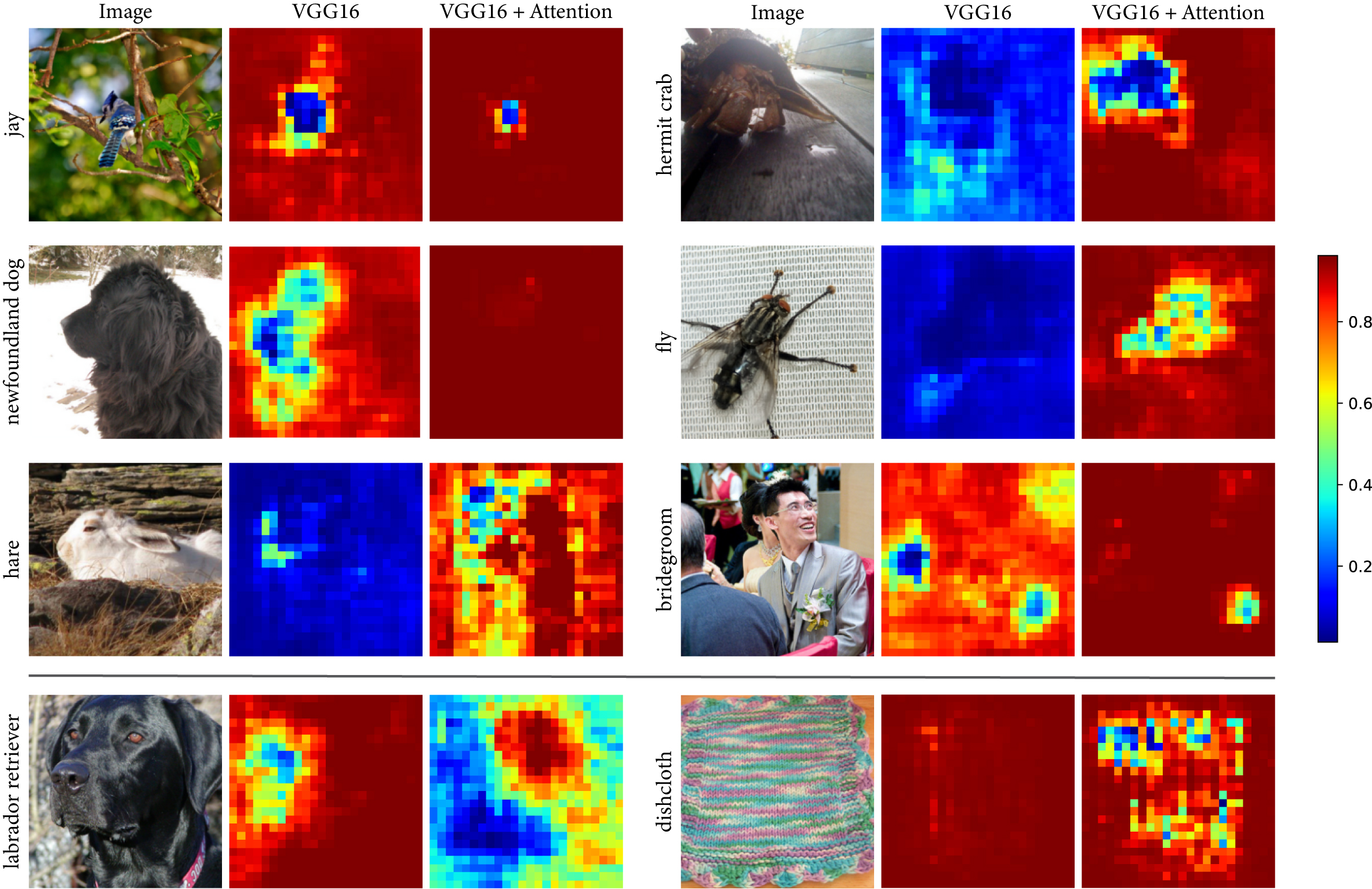}
  \caption{Example results from the occlusion experiment. Each example contains three images (from left to right): the input image, the heatmap showing the probability of correct classification by VGG16 given occlusion applied to different locations in the image, and the heatmap after adding the attention. The ImageNet class label is shown on the left. The first three rows show examples of when attention makes the model's performance less sensitive to occlusion. The last row shows examples of the opposite.}
  \label{fig:occlusion_example}
\end{figure}

To evaluate how self-attention changes the feature representation, we further perform an occlusion experiment \cite{zeiler2014visualizing}. For each image in the validation dataset, a fixed-sized window ($32 \times 32$) centered at a specific location is occluded with a grey square. The center of this occlusion is iterated throughout the whole image (stride = $8$) for individual trials of the occlusion experiment. Each trial of occlusion outputs a probability of the correct class. It is expected that after occluding different portions of the input image, the model prediction (i.e., the probability of classifying the occluded input as the correct label) may result in different confidence levels. If the occluded region includes a key feature of the correct class, the probability may drop significantly. The effect of occlusion is evaluated and visualized as a heat map, which shows the probability of correct classification as a function of the center of occlusion. For example (see Fig.~\ref{fig:occlusion_example}), VGG16 fails to classify the image of jay (a bird) as the correct label when any part of the bird is occluded. After adding the self-attention layer, the classification is compromised only when a very small part of the image is occluded. Similarly, in the image of a bridegroom, the classification performance drops only when a key feature (the Boutonnière) is occluded, whereas the performance of VGG16 is sensitive to occlusions at multiple spots. In another example image (Newfoundland dog), the attention-enhanced model is insensitive to the occlusion placed anywhere. In rarer cases, attention makes the model more sensitive to occlusion. See the last row of Fig.~\ref{fig:occlusion_example}. Such cases usually involve a large-sized object in the image and the object identity is most defined by the local texture (e.g., the dishcloth). Overall, adding the self-attention helps aggregate information across the image and makes the model much less sensitive to image occlusion. 

\begin{figure}[H]
  \centering
  \includegraphics[width=0.8\textwidth]{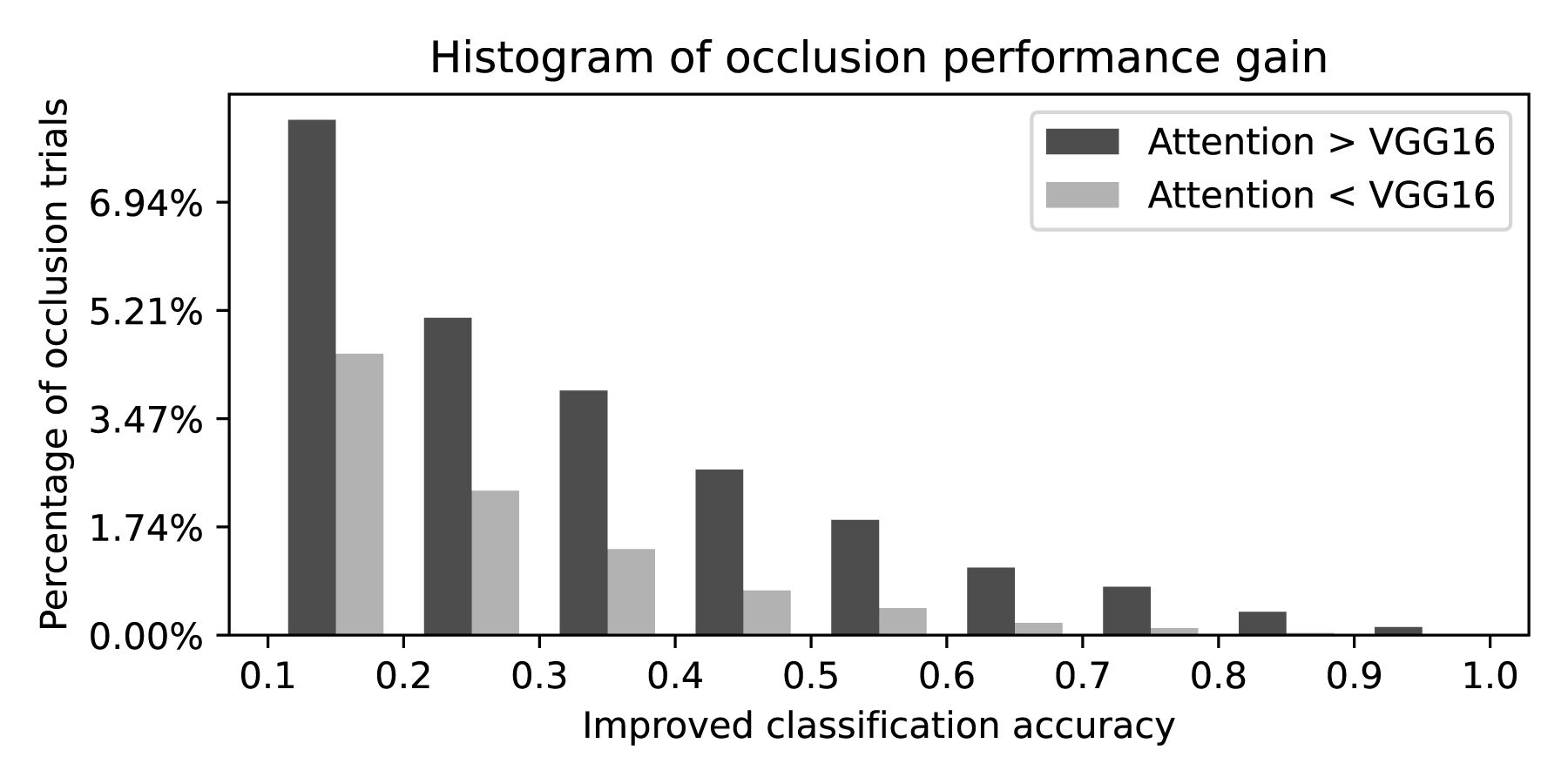}
  \caption{Quantitative results with occlusion placed at different parts of the input image in different trials. The $y$ axis shows the percentage of trials in which the attention-enhanced model shows better (dark grey) or worse (light grey) performance than VGG16. The $x$ axis shows the absolute difference in the probability of correct classification between the model with and without attention.}
  \label{fig:occlusion_stat}
\end{figure}

Quantitatively, we compare the probability of correct classification between VGG16 and its variation with attention for each trial of occlusion. We count the number of trials that the attention mechanism increases (or decreases) the probability of correct classification relative to VGG16, and evaluate the histogram by the size of increase (or decrease). As shown in Fig.~\ref{fig:occlusion_stat}, visual attention improves the classification of occluded images in many more trials than its baseline VGG16. Overall, the self-attention layer makes the model more robust against occlusion. 

\newpage
\subsection{Visual grounding of natural language with MS COCO} \label{sec:training_second_stage}

We pretrain the two-stream model with cross-modal contrastive learning on MS COCO dataset \citep{lin2014microsoft} as described in the main text Section 3.1. The training dataset consists of $118287$ images, each having $5$ captions. For each epoch, we randomly choose $1$ out of the $5$ captions. At this training stage, we freeze the convolutional layers (i.e., VGG16) in the visual stream and lower layers of the Bert encoder in the language stream.

For each self-attention layer in Bert, we also freeze the weights in query and key transformations (both are linear layers). This is motivated by two reasons. First, we want to control the number of learnable parameters to avoid over-fitting. Second, we want to separate the functional role of query ($\vb*{Q}$), key ($\vb*{K}$), and value ($\vb*{V}$) in self-attention. $\vb*{Q}$s and $\vb*{K}$s are trained to learn the syntactic and contextual relation between words in a sentence and $\vb*{V}$s are trained to learn word meanings. While $\vb*{Q}$s and $\vb*{K}$s have been trained adequately in the pretrained Bert, we freeze them to maintain the learned syntactic relations. We focus on refining $\vb*{V}$s in order to learn and represent word meanings in reference to visual perception, while leveraging both \emph{textual} context and \emph{multimodal} context.

We train the model with \texttt{Adam} optimizer (learning rate$=5\mathrm{e}{-5}$, weight decay$=5\mathrm{e}{-7}$, $\beta=(0.95, 0.999)$; dropout$=0.3$; learning rate decay by half after every $15$ epochs; batch size=$180$; total training epochs=$100$). The temperature parameter in the contrastive loss is always set to $0.1$. The parallel training is done with $3$ Nvidia GeForce RTX 2080 Ti Graphics Card (each card with $11$ GB memory).

\begin{figure}[htp]
  \centering
  \includegraphics[width=0.55\textwidth]{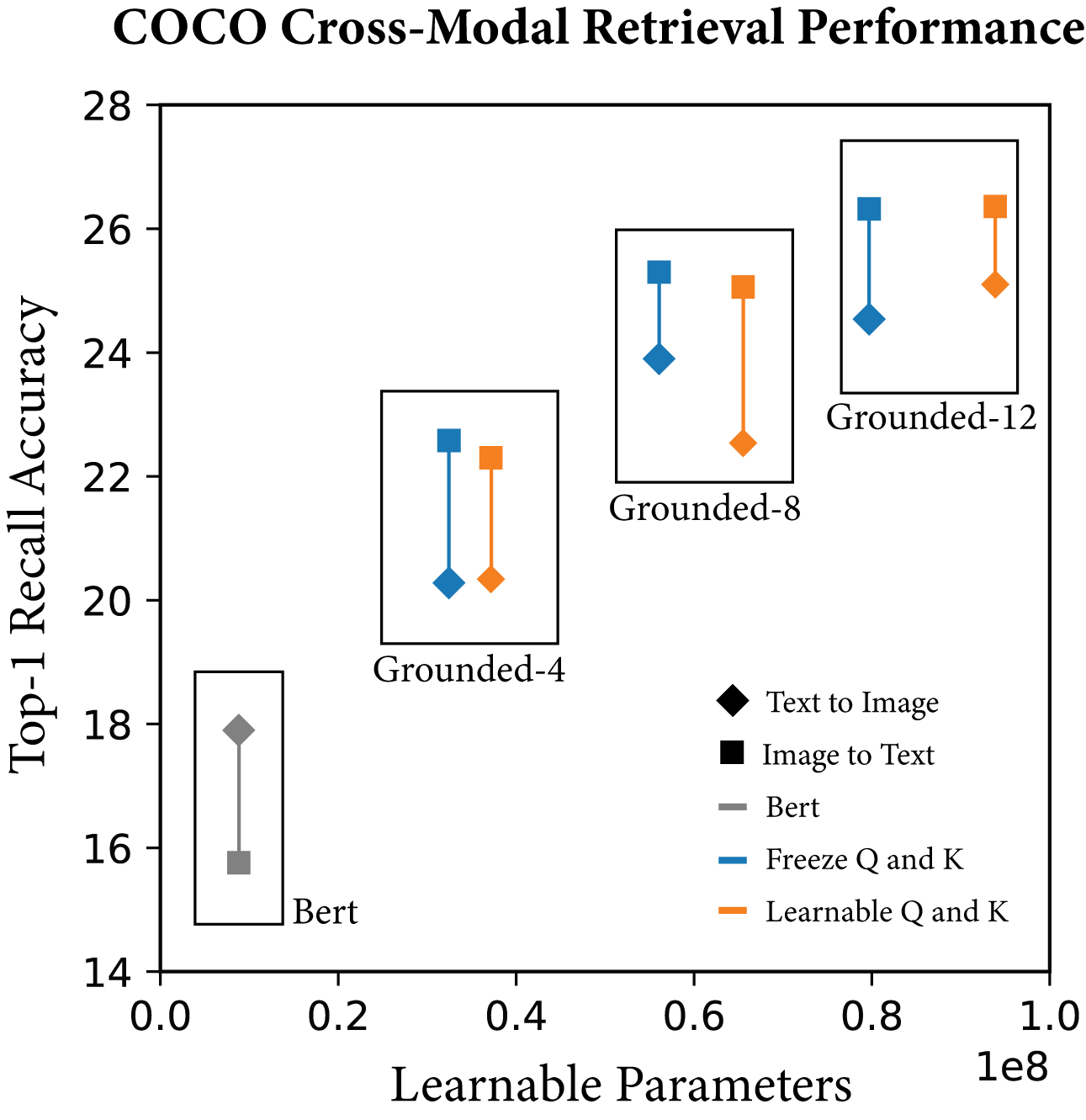}
  \caption{Cross-modal retrieval performance on MS COCO. The $x$ axis refers to the number of learnable parameters at this training stage. The label under the black box refers to a "grounded" language stream with the top $4$, $8$, or $12$ layers learnable, while the rest layers are fixed as are pretrained. \texttt{Bert}: the whole language stream is frozen; \texttt{Grounded-k}: the top $k$ layers in Bert are learnable.}
  \label{fig:COCO_retrieval_performance}
\end{figure}

Fig.~\ref{fig:COCO_retrieval_performance} shows the image-to-text and text-to-image retrieval performance on the MS COCO validation set, which contains $5000$ images. The results suggest that making more layers in Bert learnable, which is interpreted as earlier stage of visual grounding, tends to result in better cross-modal retrieval accuracy, while freezing weights on query and key transformations (blue dots) can reduce the number of learnable parameters without compromising the performance.

An ablation study on contrastive losses further suggests that both $\text{Loss}_l$ and $\text{Loss}_v$ (as defined in Equation $2$ in the main text) are important for this stage of training (Table~\ref{table:ablation_loss}). In brief, if we only use $\text{Loss}_l$ (which contrasts between positive and negative language keys), the text-to-image retrieval drops by $6.5\%$, while the image-to-text retrieval just increases by $0.2\%$. On the other hand, if we only use $\text{Loss}_v$ (which contrasts between positive and negative image keys) for training, the image-to-text retrieval significantly drops from $25.3\%$ to $1.4\%$, although the text-to-image retrieval performance increases by $1.6\%$. The overall cross-modal retrieval achieves the best performance when both losses are used for model training.

\begin{table}[htp]
  \caption{Ablation study of cross-modal contrastive losses (tested on Grounded-8 model).}
  \label{table:ablation_loss}
  \centering
  \begin{tabular}{m{3cm} m{3cm} m{3cm}}
    \toprule
     & \multicolumn{2}{c}{Cross-modal retrieval accuracy (\%)}  \\
    \cmidrule(r){2-3}
    Loss function               & image-to-text     & text-to-image  \\
    \midrule
    $\text{Loss}_l + \text{Loss}_v$              & 25.3   & 23.9  \\
    $\text{Loss}_l$              & 25.5   & 17.4  \\
    $\text{Loss}_v$              & 1.4   & 25.5  \\
    \bottomrule
  \end{tabular}
\end{table}

\newpage
\subsection{Visual grounding of object relations with Visual Genome}

\subsubsection{Details in cross-modal attention and bilinear relational modules}
For each head in the cross-modal attention module, the queries (Query$_L$) are from the object descriptions encoded by the language stream; the keys (Key$_V$) are from visual features in the visual stream (Eq.~\ref{eq:KQV_cross_attention}). The attention score $\vb*{A}_{L \to V}$ is the inner-product between Query$_L$ and Key$_V$ (Eq.~\ref{eq:cross_attention_score}). The attention-weighted sum of the Value$_V$ from the visual stream (Eq.~\ref{eq:KQV_cross_attention}) is concatenated across different heads to generate a visually grounded object representation (Eq.~\ref{eq:grounded_object}).

\begin{equation}
    \text{Key}_V^i = \vb*{V} \vb*{W}_{K}^i, \quad
    \text{Value}_V^i = \vb*{V} \vb*{W}_{V}^i, \quad
    \text{Query}_L^i = \vb*{L} \vb*{W}_{Q}^i
    \label{eq:KQV_cross_attention}
\end{equation}

\begin{equation}
    \vb*{A}_{L \to V}^i =  \text{softmax} \{ \text{Query}_L^i (\text{Key}_V^i) ^T \ / \sqrt{d} \}, \quad
    \label{eq:cross_attention_score}
\end{equation}

\begin{equation}
    \vb*{O} = \text{concat} \{ \vb*{A}_{L \to V}^1 \text{Value}_V^1, \cdots, \vb*{A}_{L \to V}^h \text{Value}_V^h \}
    \label{eq:grounded_object}
\end{equation}

where $i$ in Eq.~\ref{eq:KQV_cross_attention} and Eq.~\ref{eq:cross_attention_score} refers to the $i$-th attention head. $d$ in Eq.~\ref{eq:cross_attention_score} refers to the query/key feature dimension. $h$ in Eq.~\ref{eq:grounded_object} refers to the total number of attention heads (by default $h=8$).

In the bilinear relational module, each predicate is represented by a matrix $\vb*{R}_p \in \mathbb{R}^{d \times d}$ that encodes a specific relation between two visual objects. The relation score defined in Eq.~\ref{eq:bilinear_operator} measures to how well the predicate $p$ describes the relation between a subject $s$ and a object $o$. We constrain the Frobenius norm (Eq.~\ref{eq:optimal_relation}) to each relational embedding matrix such that the compositionality of relations follows the transitivity property. That is, the representation of the optimal relation between the object \texttt{1} and object \texttt{3} is the multiplication of the relation between object \texttt{1} and object \texttt{2} and the relation between object \texttt{2} and object \texttt{3} (Eq.~\ref{eq:transitivity}):

\begin{equation}
    \small
    S(\vb*{r}_s, \vb*{r}_o; \vb*{R}_p) = F_p(\vb*{r}_s, \vb*{r}_o) = \vb*{r}_s \vb*{R}_p \vb*{r}_o^T
    \label{eq:bilinear_operator}
\end{equation}

\begin{equation}
    \vb*{R}_{(\vb*{r}_1, \vb*{r}_2)}^* = \argmax _{\| \vb*{R} \|_F = 1} S(\vb*{r}_1, \vb*{r}_2; \vb*{R}) = \frac{\vb*{r}_1^T \vb*{r}_2}{\|\vb*{r}_1\|_2\|\vb*{r}_2\|_2}
    \label{eq:optimal_relation}
\end{equation}

\begin{equation}
    \vb*{R}_{(\vb*{r}_1, \vb*{r}_3)}^* = \vb*{R}_{(\vb*{r}_1, \vb*{r}_2)}^* \vb*{R}_{(\vb*{r}_2, \vb*{r}_3)}^*
    \label{eq:transitivity}
\end{equation}

\subsubsection{Data cleaning for relation and object labels and training parameters.}

Since the relation labels are imbalanced in the original Visual Genome dataset \citep{krishna2017visual}, we filter the data to create a cleaner dataset for training and testing the proposed model to perform visual relation prediction task as well as the object classification, which is used as an auxiliary learning objective. To define object labels, we first extract the WordNet \citep{miller1995wordnet} synset for each object in Visual Genome data annotations. We investigate the distribution of the hypernyms of all object synsets and summarize them into $55$ general classes as shown in Table~\ref{table:object_labels}. To define relation labels, we first extract the "predicate" term for each pair of objects in the data annotations and only preserve the ones with more than $250$ instances in the Visual Genome dataset (which remains $292$ out of $37342$ unique labels). We then manually merge equivalent predicates into a single relation label (e.g., merge “\emph{near}, \emph{next to}, \emph{on side of}, \emph{beside}, \emph{standing next to}, \emph{next}, \emph{standing near}, \emph{to right of}, \emph{near a}, \emph{close to}, \emph{on side}, etc.” to “\textbf{near}”). In this way, we end up with $114$ unique relation labels as shown in Table~\ref{table:relation_labels}. Furthermore, we remove image samples with fewer than 5 subject-predicate-object triplets and keep a total of $98512$ images. We then randomly split this cleaned dataset into training ($93512$ samples) and testing ($5000$ samples) set. 

At this stage of training, we also freeze the CNN (the VGG16 encoder) in the visual stream and lower layers in the language stream, only keeping top $2$ layers in Bert learnable. We train the model with \texttt{Adam} optimizer (learning rate$=1\mathrm{e}{-5}$, weight decay$=5\mathrm{e}{-7}$, $\beta=(0.95, 0.999)$; dropout$=0.1$; learning rate decay by half after every $15$ epochs; batch size=$180$; total training epochs=$150$). The temperature parameter in the contrastive loss is always set to $1.0$. The parallel training is done with $3$ Nvidia GeForce RTX 2080 Ti Graphics Card (each card with $11$ GB memory).

\begin{table}[htp]
  \small
  \caption{Object labels (defined by WordNet synsets) for visually grounded object classification.}
  \label{table:object_labels}
  \centering
  \begin{tabular}{m{4.2cm} | m{4.2cm} | m{4.2cm}}
    \toprule
    feline.n.0                  & equine.n.01               & mammal.n.01       \\
    bird.n.01                   & animal.n.01               & body\_part.n.01   \\
    bread.n.01                  & vegetable.n.01            & fruit.n.01        \\
    meat.n.01                   & beverage.n.01             & food.n.01         \\
    tree.n.01                   & herb.n.01                 & vessel.n.02       \\ 
    wheeled\_vehicle.n.01       & aircraft.n.01             & vehicle.n.01      \\
    road.n.01                   & clothing.n.01             & furniture.n.01    \\
    tableware.n.01              & home\_appliance.n.01      & stairs.n.01       \\
    building\_material.n.01     & decoration.n.01           & room.n.01         \\
    building.n.01               & container.n.01            & surface.n.01      \\
    machine.n.01                & measuring\_instrument.n.01 & instrument.n.01  \\
    tool.n.01                   & device.n.01               & paper.n.01        \\
    man.n.01                    & woman.n.01                & person.n.01       \\
    equipment.n.01              & sport.n.01                & activity.n.01     \\
    symbol.n.01                 & sign.n.02                 & number.n.02       \\ 
    writing.n.02                & body\_of\_water.n.01      & facility.n.01     \\
    geological\_formation.n.01  & location.n.01             & atmospheric\_phenomenon.n.01 \\
    phenomenon.n.01             & communication.n.02        & structure.n.01    \\
    artifact.n.01 &&\\
    \bottomrule
  \end{tabular}
\end{table}

\begin{table}[htp]
  \small
  \caption{Relation labels for visual relation prediction task.}
  \label{table:relation_labels}
  \centering
  \begin{tabular}{m{2.1cm} | m{2.1cm} | m{2.1cm} | m{2.1cm} | m{2.1cm}}
    \toprule
    on          & have          & in        & of            & wear  \\
    with        & behind        & hold      & near          & under \\
    by          & above         & sit       & in front of   & to    \\
    at          & over          & for       & around        & ride  \\
    stand       & hang          & carry     & eat           & walk  \\
    cover       & play          & lay       & along         & among \\
    and         & watch         & belong to & painted       & against   \\
    from        & parked        & made of   & say           & covered   \\
    mounted     & across        & fly       & lying         & grow  \\
    use         & outside       & cross     & worn          & printed   \\
    full of     & filled with   & swing     & built         & pull  \\
    touch       & adorn         & a         & hit           & support   \\
    written     & lean          & drive     & rest on       & held  \\
    connected to    & cut       & throw     & line          & through   \\
    float       & show          & face      & graze         & cast  \\
    stick out of    & catch       & drink     & reflected in  & be    \\
    beyond      & lead          & read      & swim          & white \\
    off         & seen          & push      & shining on    & ski   \\
    wait        & surf          & down      & make          & feed  \\
    run         & take          & enjoy     & that          & at end of \\
    stuck       & reflect       & stacked   & black         & plugged   \\
    overlook    & form          & without   & do            & kick  \\
    visible on  & brush         & blue      & work on       & \\
    \bottomrule
  \end{tabular}
\end{table}

\newpage
\subsubsection{Ablation study on loss functions}

\begin{figure}[htp]
  \centering
  \includegraphics[width=0.32\textwidth]{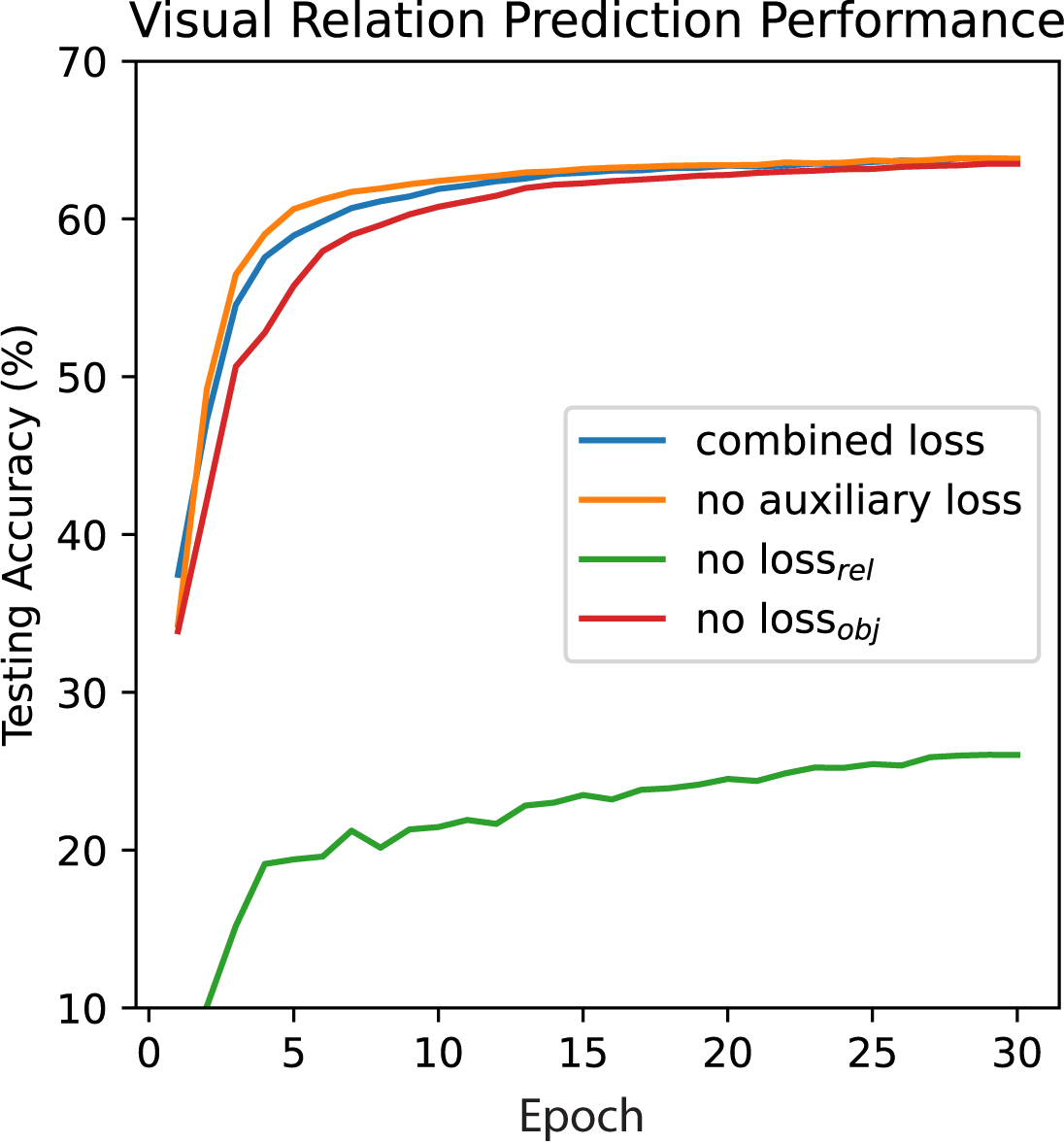}
  \caption{Learning curves for visual relation prediction on the testing dataset for the first $30$ training epochs. The blue curve shows the performance when the model is trained with the loss that combines $\text{Loss}_{\text{rel}}, \text{Loss}_{\text{obj}}$, and the auxiliary loss for object classification, which is the default setting as mentioned in the main text of this work. The other three curves show the performance when the model is trained by excluding one of the three losses, as indicated in the figure legend.}
  \label{fig:VG_testing_accuracy}
\end{figure}

\begin{figure}[htp]
  \centering
  \includegraphics[width=\textwidth]{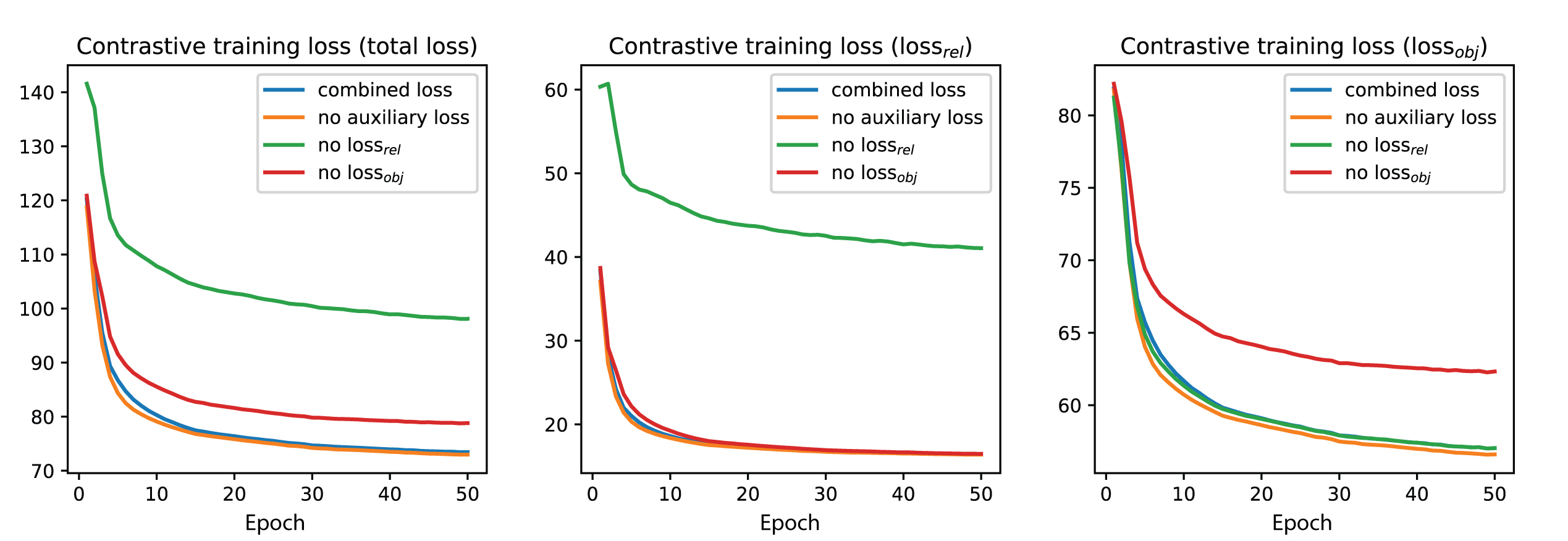}
  \caption{Learning curve of the contrastive loss functions. The \emph{total loss} in the first figure refers to the summation $\text{Loss}_{\text{rel}}+\text{Loss}_{\text{obj}}$.}
  \label{fig:VG_ablation_loss}
\end{figure}

Since the learning objective includes three loss functions ($\text{Loss}_{\text{rel}}, \text{Loss}_{\text{obj}}$, the auxiliary loss for object classification as described in Section 3.2). We also check how each loss function contributes to the model performance. We  perform an ablation experiment by excluding out one loss at a time. The results suggest that the model has the best performance on relation prediction by combining all three losses (Table~\ref{fig:VG_ablation_loss}, Fig.~\ref{fig:VG_testing_accuracy}). Whereas other losses do not appear to make a major difference, if $\text{Loss}_{\text{rel}}$ is excluded, the model would have much worse performance. This result suggests that $\text{Loss}_{\text{rel}}$ is the key component that allows the model to learn visual relation. We also find that $\text{Loss}_{\text{obj}}$ and $\text{Loss}_{\text{rel}}$ are somewhat entangled (i.e., minimizing one loss tends to decrease the other loss, see Fig.~\ref{fig:VG_ablation_loss}), but the model learns faster (Fig.~\ref{fig:VG_testing_accuracy}) if we combine both two contrastive losses.

\begin{table}[htp]
  \caption{Testing performance on ablation study of loss functions.}
  \label{table:VG_ablation_loss}
  \centering
  \begin{tabular}{m{3cm} m{2.2cm} m{2.2cm} m{2.2cm}}
    \toprule
    Model               & Object Classification     & Relation Prediction (Top-1)     & Relation Prediction (Top-10) \\
    \midrule
    Combined loss       & 97.71     & \textbf{64.26}     & \textbf{95.21} \\
    No auxiliary loss   & 0.60      & 64.19     & 94.99 \\
    No $\text{Loss}_{\text{rel}}$ & 97.95     & 29.97     & 68.78 \\
    No $\text{Loss}_{\text{obj}}$ & \textbf{98.39}     & 64.14     & 95.14 \\
    \bottomrule
  \end{tabular}
\end{table}

\newpage
\subsubsection{Examples on testing dataset}
\begin{figure}[H]
  \centering
  \includegraphics[width=\textwidth]{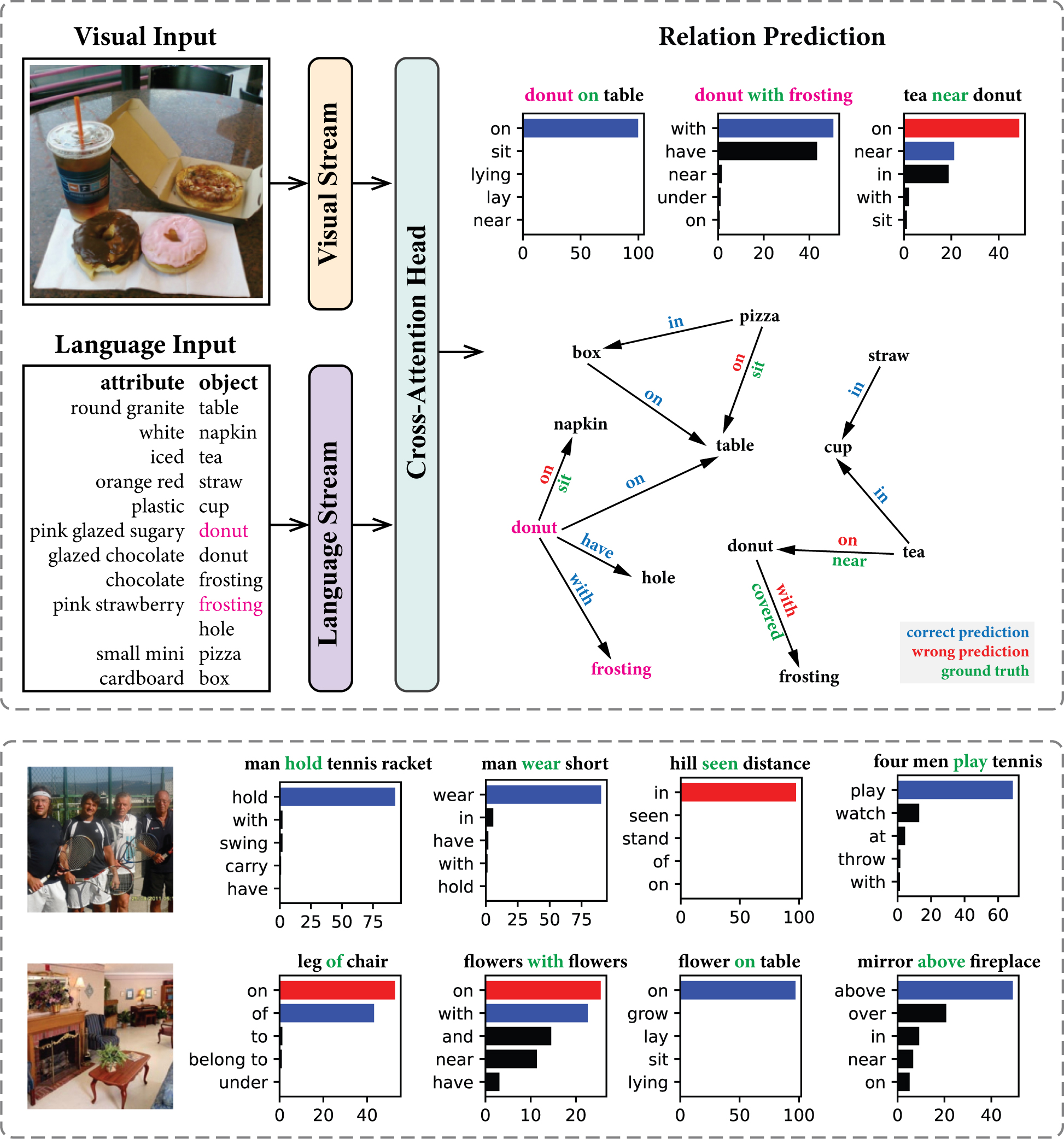}
  \caption{Examples of visual relation prediction task performance on testing dataset. Top: For visual grounding of object relations, the visual input is a natural image and the language input is a set of object descriptions. The bar charts show the examples for top-$5$ predicted visual relation of a paired objects. The directed graph shows top-$1$ predicted visual relations on all object pairs with ground truth labels in this example. Bottom: Other examples in the testing dataset.}
  \label{fig:visual_relation_prediction}
\end{figure}

\newpage
\section{Evaluating the effect of visual grounding on the language stream}

We first extract the word embeddings from the language model, which always has a Bert-base structure but with different levels of visual grounding (\textbf{Bert}: no visual grounding; \textbf{Grounded}: visual grounding of natural language; \textbf{Relational Grounded}: visual grounding of object relations). To do this, we input every single word (or phrase) preceded with a special token \texttt{[CLS]} and followed by a special token \texttt{[SEP]} (according to the original Bert \citep{devlin2019bert} paper) into the language stream, and use the average pooled output from the last hidden layer as the extracted word embedding. Since a few output feature channels have much larger standard deviations than other feature channels, we further use the mean and standard deviation of the output representation of the $30,522$-token vocabulary (which defines the embedding layer in Bert \citep{wu2016google}) to standardize each single word representation. The same process is applied to both the Bert model and the visually grounded language models. For each input word (or phrase), its output embedding is a $d$-dimensional vector ($d=768$). All embeddings of commonly used English words from the vocabulary set $S$ (defined by the SemCat dataset \citep{csenel2018semantic}) form a set of vector representations in this high-dimensional semantic space.

\subsection{Extended results on principal component analysis}

\begin{figure}[H]
  \centering
  \includegraphics[width=\textwidth]{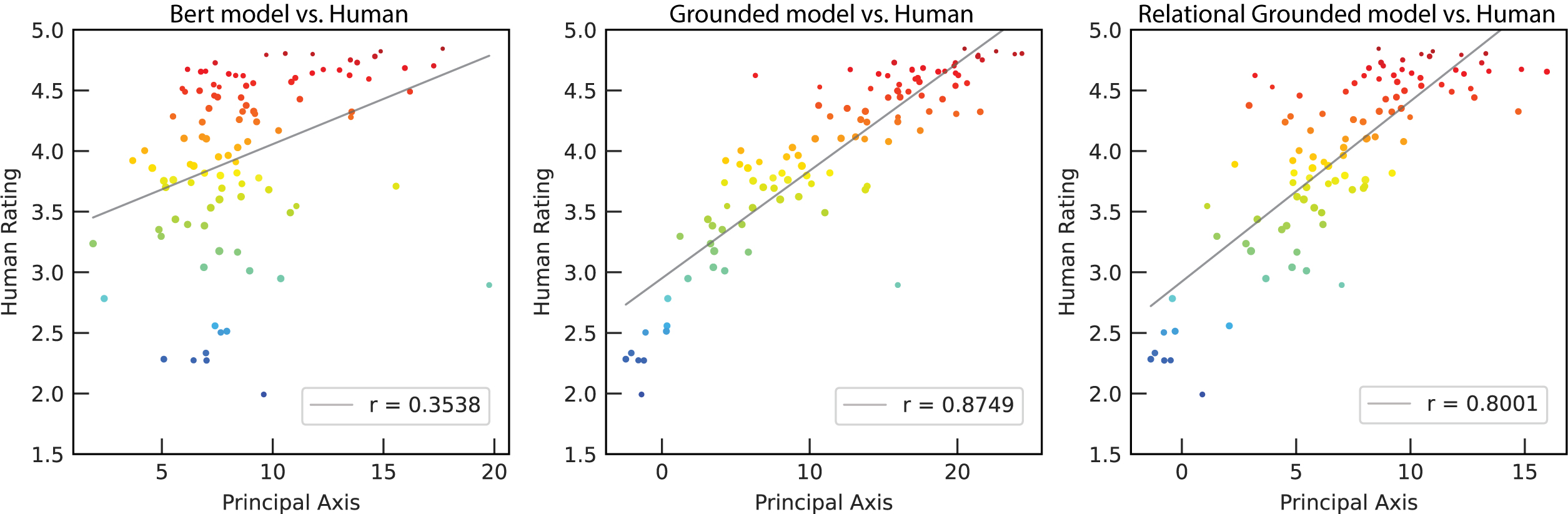}
  \caption{The first principal component in word representation space captures concrete-abstract axis only after visual grounding. The color-coding of each dot indicates the averaged human-rated concreteness score of a category (blue: abstract; red: concrete).}
  \label{fig:concrete_abstract_comparison}
\end{figure}

\begin{table}[H]
  \caption{Correlation between the $1^{\text{st}}$ principal axis and human-rated word concreteness}
  \label{table:concreteness}
  \centering
  \begin{tabular}{m{2cm} m{1.5cm} m{1.5cm} m{1.5cm}}
    \toprule
     & \multicolumn{3}{c}{Correlation (Pearson's r)}  \\
    \cmidrule(r){2-4}
    Group           & Bert      & Grounded      & \small{Relational Grounded} \\
    \midrule
    word-level      & 0.1040    & 0.6615        & \textbf{0.6948} \\
    category-level  & 0.3538    & \textbf{0.8749}        & 0.8001 \\
    \bottomrule
  \end{tabular}
\end{table}

\begin{figure}[H]
  \centering
  \includegraphics[width=\textwidth]{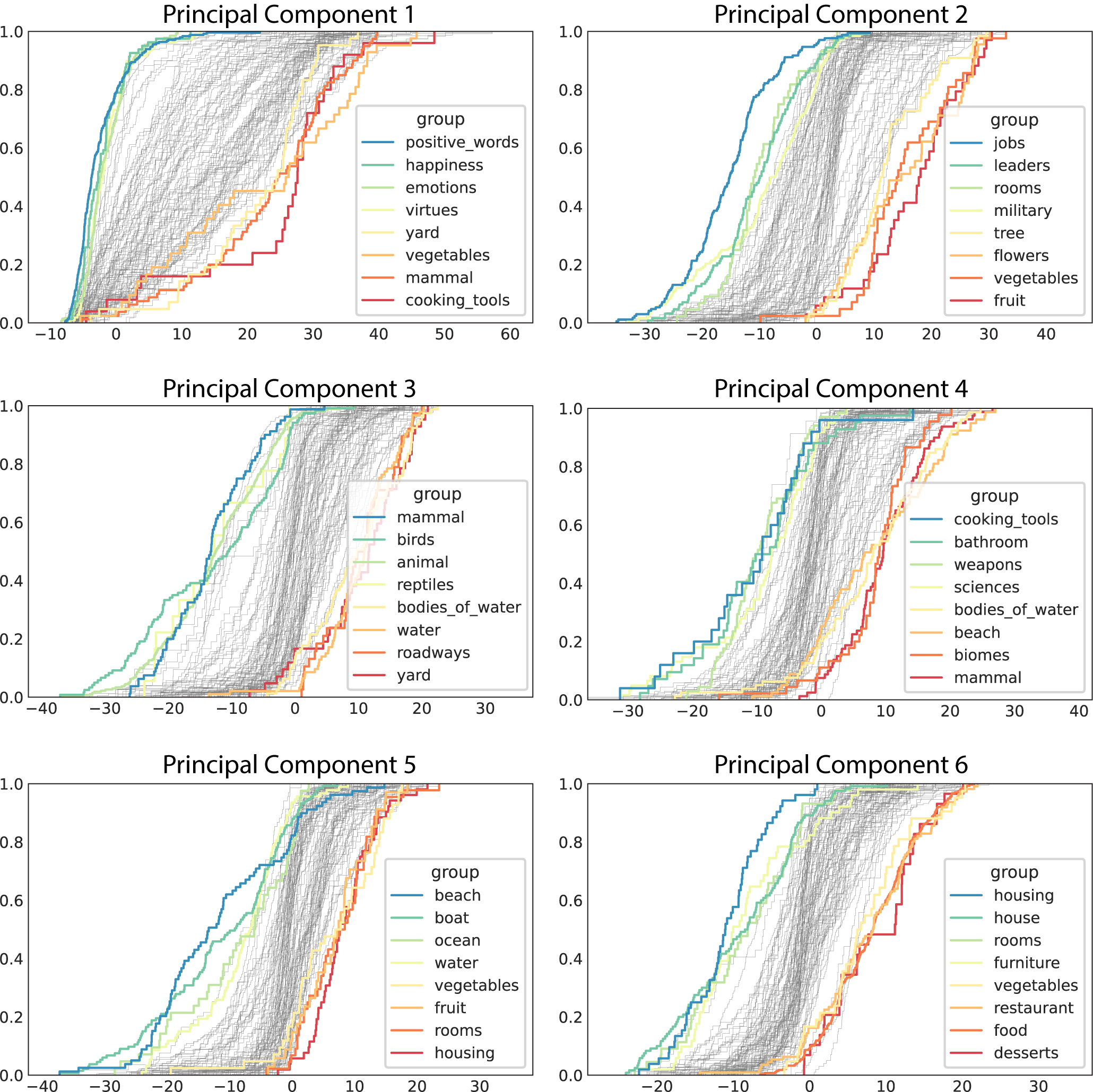}
  \caption{Other principal components in the visually grounded word representation space. Each plot shows a set of cumulative distribution functions (CDFs) for different word categories after projecting to a principal component. The resulting principal dimensions capture the semantic attributes that can be interpreted by human intuition. PC1: abstract vs. concrete; PC2: human vs. non-human; PC3: object vs. scene; PC4: artificial vs. natural; PC5: outdoor vs. indoor; PC6: non-food vs. food.}
  \label{fig:PC_1_to_6}
\end{figure}

\newpage
\subsubsection{2D visualization of the first three principal axes}

\begin{figure}[H]
  \centering
  \includegraphics[width=\textwidth]{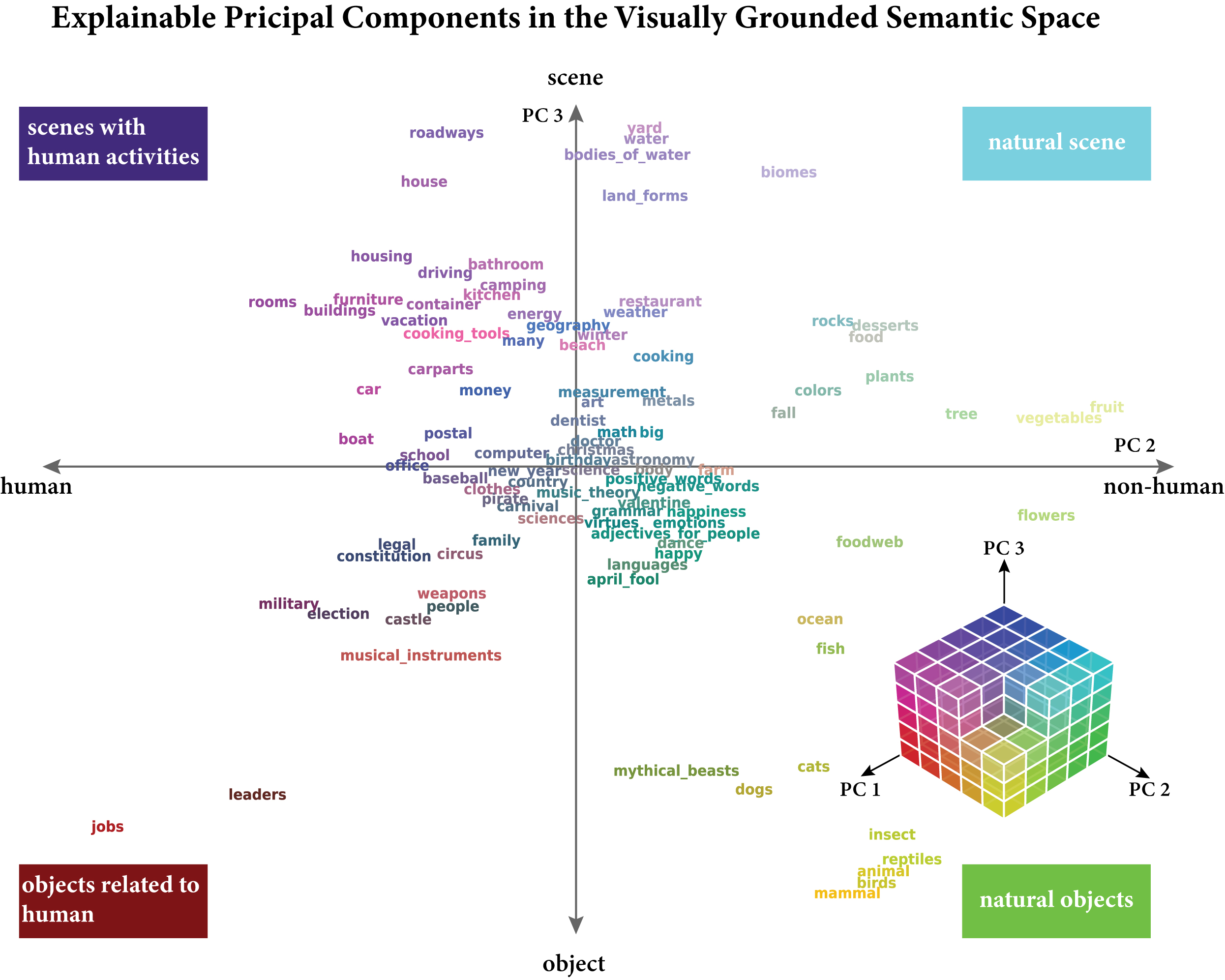}
  \caption{2D visualization of PC2 and PC3.}
  \label{fig:2d_pc_2_3}
\end{figure}

\begin{figure}[H]
  \centering
  \includegraphics[width=\textwidth]{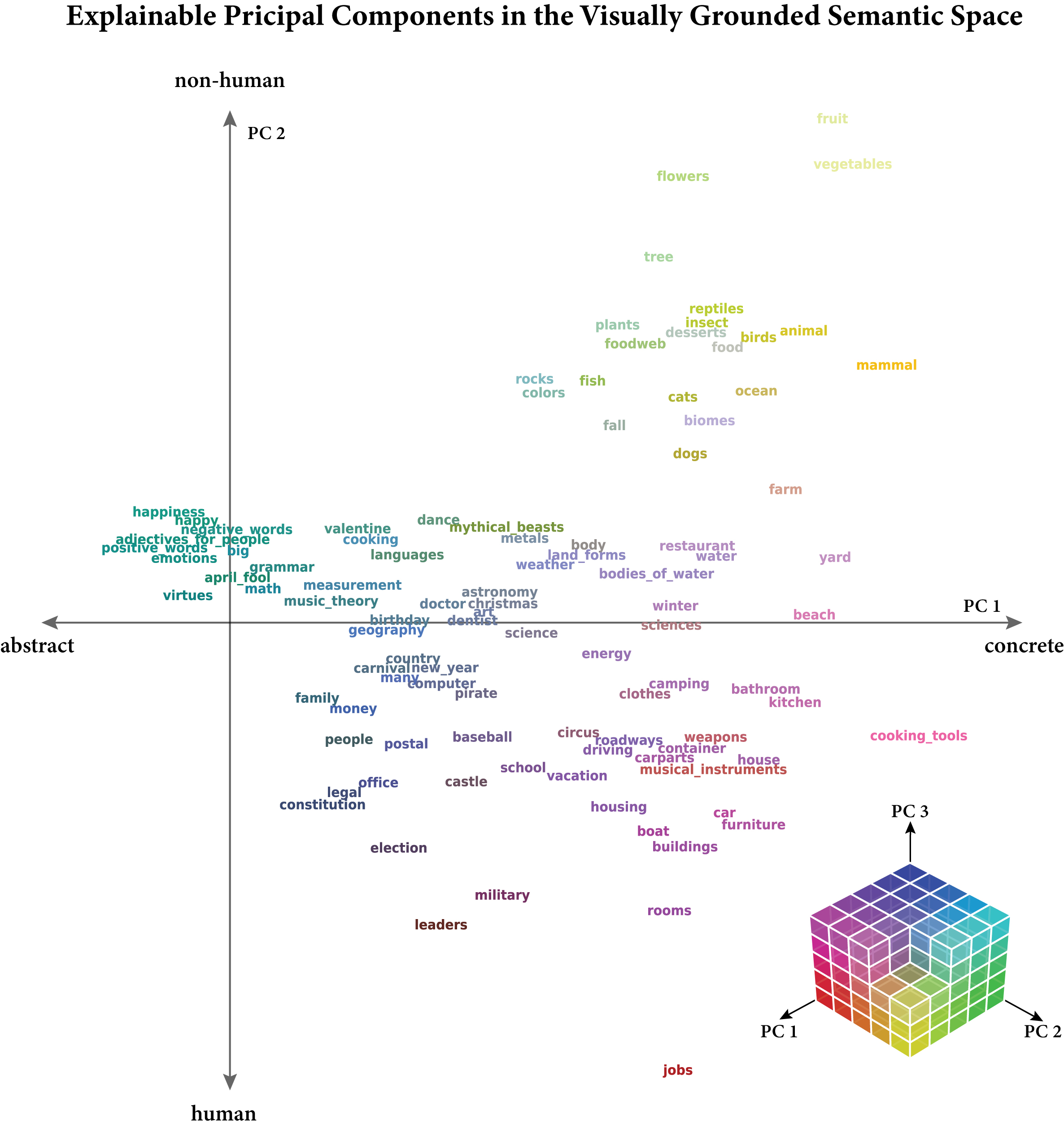}
  \caption{2D visualization of PC1 and PC2.}
  \label{fig:2d_pc_1_2}
\end{figure}

\begin{figure}[H]
  \centering
  \includegraphics[width=\textwidth]{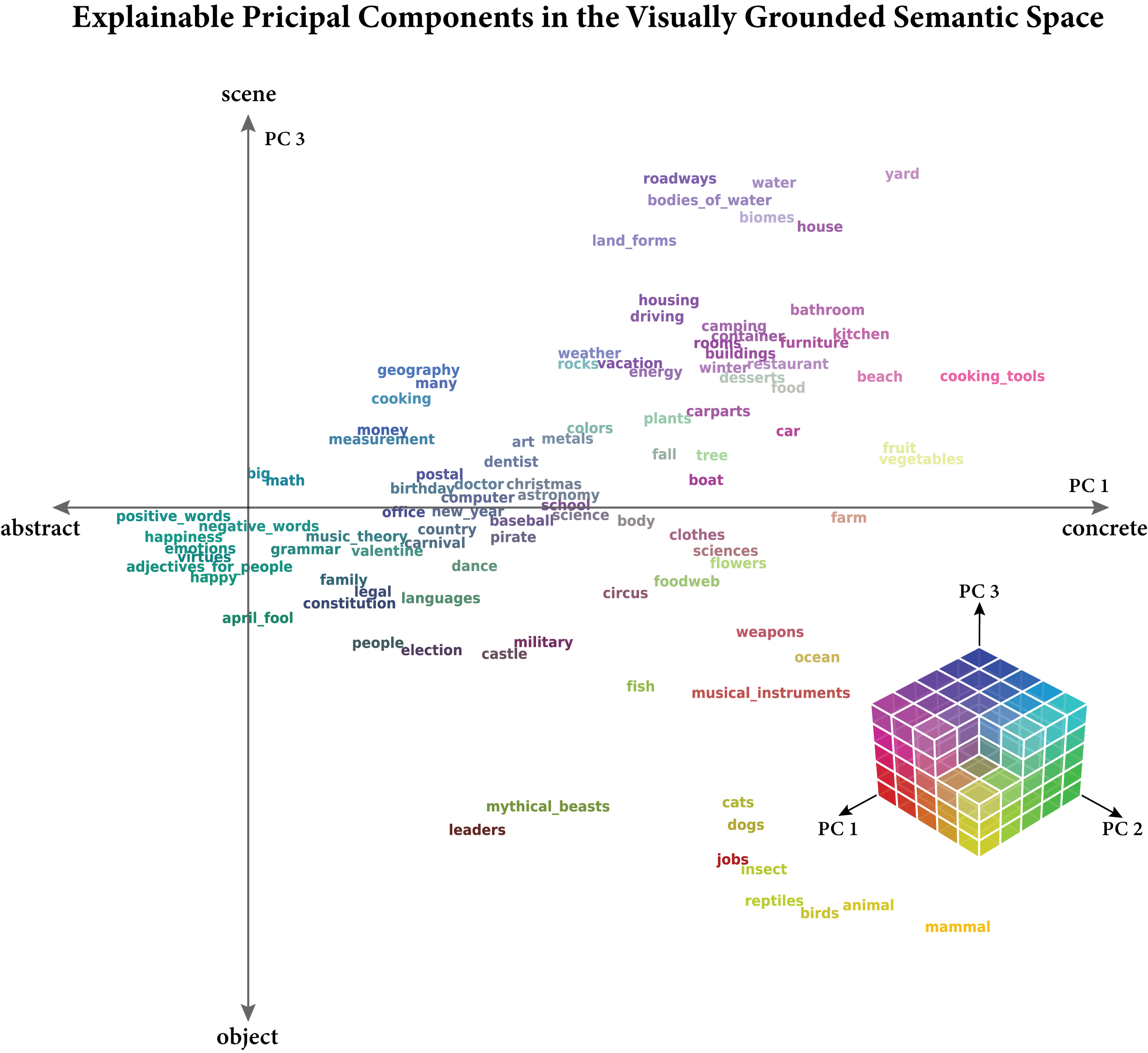}
  \caption{2D visualization of PC1 and PC3.}
  \label{fig:2d_pc_1_3}
\end{figure}

To facilitate the understanding of how conceptual representations are organized in the grounded semantic space, we visualize the $100$ word categories in a linear subspace spanned by the first three principal axes. We first color-code each word category by using an RGB code: (PC1, red), (PC2, green), (PC3, blue). Only for this color-coding purpose, the coefficients of each principal component are linearly re-scaled into the range $[0,1]$. We then project the three dimensional representations of the $100$ word categories further into three 2D planes, as shown in Fig.~\ref{fig:2d_pc_2_3} (PC2 vs. PC3), Fig.~\ref{fig:2d_pc_1_2} (PC1 vs. PC2), and Fig.~\ref{fig:2d_pc_1_3} (PC1 vs. PC3).

The result of PC 2 vs. PC 3 (Fig. \ref{fig:2d_pc_2_3}) suggests that the first quadrant represents concepts describing natural scenes (e.g. biomes, rocks), the second quadrant represents concepts related to scenes with human activities (e.g. roadways, rooms), the third quadrant encodes human-related non-scene concepts (e.g. jobs, musical instruments), the fourth quadrant encodes non-human objects (e.g. animal, foodweb). The abstract words (e.g. emotions, happiness) are squeezed around the origin in this projected representation.

In the 2D projection of PC1 vs. PC2 (Fig.~\ref{fig:2d_pc_1_2}) or PC1 and PC3 (Fig.~\ref{fig:2d_pc_1_3}), we also observe that although concrete concepts are distributed and scattered widely in the semantic space, the abstract concepts tend to be squeezed around the origin. This is perhaps caused by the lack of information supporting rich representations for emotional words since we only grounded the language model in vision.

\newpage
\subsection{Extended results on semantic norm prediction}

To investigate whether the visually grounded word embeddings capture semantic norms defined by humans, we train a logistic regression model with L1 regularization from the word embeddings to predict each binary semantic feature defined in the CSLB dataset \citep{devereux2014centre}.

Since many binary semantic norms in the CSLB dataset contain very few positive word samples, we first filter out the ones with fewer than $5$ positive samples, which results in $390$ out of $2725$ feature norms, dividing into five feature types: $156$ “visual perceptual” features (e.g. \texttt{has\_wheels}); $29$ “other perceptual” features (e.g. \texttt{has\_flavors}); $94$ “functional” features (e.g. \texttt{does\_cut}); $65$ “encyclopaedic” features (e.g. \texttt{is\_dangerous});  $46$ “taxonomic” features (e.g. \texttt{is\_clothing}). For the $i$-th semantic norm, we build a binary classifier with a logistic regression model $p^i$ (\citep{li2017distributional}):

\begin{equation}
    p^i (y_{ij}=1|\vb*{x_j}) = \sigma(\vb*{w_i}^T\vb*{x_j}),
    \label{eq:logistic_regression}
\end{equation}

Here $\vb*{x_j}$ is the word representation of the $j$-th word $x_j$ after projecting onto principal axes. $\vb*{w}_i$ is a linear weight specific to the $i$-th semantic norm. $y_{ij} \in \{0,1\}$ is the binary label indicating whether the $x_j$ holds the $i$-th semantic norm. To avoid over-fitting, we add an L1-norm as a sparsity constraint. 

\begin{equation}
    \vb*{w_i}^* = \argmin _{\vb*{w_i}} \Big( -\sum_j \big[ y_{ij} \log (\sigma(\vb*{w_i}^T\vb*{x_j})) + (1-y_{ij}) \log (1-\sigma(\vb*{w_i}^T\vb*{x_j})) \big] + \lambda_i\| \vb*{w_i} \|_1 \Big)
    \label{eq:optimize_logistic}
\end{equation}

The regularization parameter $\lambda_i$ is determined by a leave-one-out cross-validation to minimize the following objective function:

\begin{equation}
    \mathcal{L}_i(\lambda_i) = \sum_j \mathcal{L}_{ij}(\lambda_i)
    \label{eq:logistic_loss}
\end{equation}

Suppose $P_i = \{ k | y_{ik}=1 \}$ and $N_i = \{ k | y_{ik}=0 \}$ are the sets consisting of positive and negative word samples for the $i$-th semantic feature, respectively ($|P_i \cup N_i|=638$, i.e., the total number of words in the CSLB dataset). 

\begin{equation}
    \small
    \mathcal{L}_{ij}(\lambda_i) = \frac{1}{ | P_i | } \sum_{k \in P_i, k \neq j} (\log p^i_{\lambda_i, j} (y_{ik}=1|\vb*{x_k})) + \frac{1}{| N_i |} \sum_{k \in N_i, k \neq j}(\log p^i_{\lambda_i, j} (y_{ik}=0|\vb*{x_k}))
    \label{eq:logistic_cross_validation}
\end{equation}

$j$ indicates the left-out word sample, and $p^i_{\lambda_i, j}$ is the trained regression model with regularization parameter $\lambda_i$. After $\lambda_i$ is determined, we train the L1-normed logistic regression model $p^i$ with all word samples and calculate the F$1$-score:

\begin{equation}
    \label{eq:semantic_norm_F1}
    \text{F}1 = \frac{\texttt{tp}}{\texttt{tp}+\frac{1}{2}(\texttt{fp}+\texttt{fn})}
\end{equation}

where \texttt{tp} refers the number of true positive cases, \texttt{fp} refers the number of false positive cases, \texttt{fn} refers the number of false negative cases. We then pairwisely compare the F$1$-score for each semantic norm across language models with different levels of visual grounding and test the statistical significance with a one-sided Wilcoxon Signed Rank Test (as shown in the main Fig. 4). Since we add a strong regularization term to avoid over-fitting, the results suggest $230$ out of $390$ semantic norms are not predictable by the ungrounded Bert model, while only $143$ and $129$ semantic norms are not predictable by the Grounded model and the Relational Grounded model respectively.

To better understand the details of this dataset and the corresponding results, we listed the top-5 semantic norms that became better predictable (according to the F1-score) after visual grounding in Table~\ref{table:semantic_norm}.

\newpage
\begin{table}[H]
  \small
  \caption{Top-5 semantic norms that are better predictable after visual grounding.}
  \label{table:semantic_norm}
  \centering
  \begin{tabular}{m{3cm} | m{4cm} | m{4cm}}
    \toprule
    Feature type        & Grounded model      & Relational Grounded model \\
    \midrule
    visual perceptual   & \texttt{has\_wheels, has\_a\_handle\_handles, has\_skin\_peel, has\_pages, has\_a\_back} & \texttt{has\_a\_picture\_pictures, has\_pages, has\_a\_barrel, has\_skin\_peel, has\_a\_seat\_seats} \\
    \midrule
    other perceptual    & \texttt{is\_heavy, is\_warm, does\_smell\_good\_nice, is\_juicy, has\_flavours}   & \texttt{is\_warm, is\_heavy, has\_flavours, is\_juicy, does\_smell\_good\_nice} \\
    \midrule
    functional          & \texttt{does\_fly, does\_contain\_hold, does\_store, does\_heat, is\_used\_to\_see} & \texttt{does\_fly, does\_heat, does\_cut, is\_used\_in\_cooking, does\_contain\_hold} \\
    \midrule
    encyclopaedic       & \texttt{is\_dangerous, is\_found\_in\_seas, has\_information, is\_healthy, does\_grow\_on\_trees} & \texttt{is\_dangerous, has\_information, does\_grow\_on\_trees, is\_found\_in\_seas, is\_found\_in\_kitchens} \\
    \midrule
    taxonomic           & \texttt{is\_clothing, is\_a\_weapon, is\_a\_vehicle, is\_a\_vegetable, is\_transport} & \texttt{is\_clothing, is\_a\_vehicle, is\_a\_vegetable, is\_medicine, is\_a\_container} \\
    \bottomrule
  \end{tabular}
\end{table}

\newpage
\subsection{Extended results on word categorization}

\subsubsection{The modified version of Sihlouette coefficient}

Suppose $N$ is the number of word categories, $\vb*{x_i}$ is the embedding of word $i$ which belongs to the category $C_i$, let

\begin{equation}
    \small
    a(i) = \frac{1}{|C_i|-1} \sum _{j \in C_i, j \neq i} d(\vb*{x_i}, \vb*{x_j}),
    \label{eq:silhouette_coefficient_a}
\end{equation}

\begin{equation}
    \small
    b(i) = \frac{1}{N-1} \sum_{k \neq i} \frac{1}{|C_k|} \sum _{j \in C_k} d(\vb*{x_i}, \vb*{x_j}),
    \label{eq:silhouette_coefficient_b}
\end{equation}

\begin{equation}
    \small
    s(i) = \frac{b(i)-a(i)}{\max {(a(i), b(i))}}
    \label{eq:silhouette_coefficient_s}
\end{equation}

where the distance metric $d$ between word embeddings is measured as the cosine distance: $d(\vb*{x_i}, \vb*{x_j}) = 1-\cos(\vb*{x_i}, \vb*{x_j})$. Here we used average instead of minimum when calculating the denominator in $b(i)$ (Eq.~\ref{eq:silhouette_coefficient_b}) since these $100$ word categories are not mutually exclusive (e.g. \texttt{mammal}, \texttt{bird}, \texttt{fish} are overlapped with \texttt{animal} in SemCat dataset).

\subsubsection{Word categorization performance compared across different training settings}

\begin{figure}[H]
  \centering
  \includegraphics[width=0.9\textwidth]{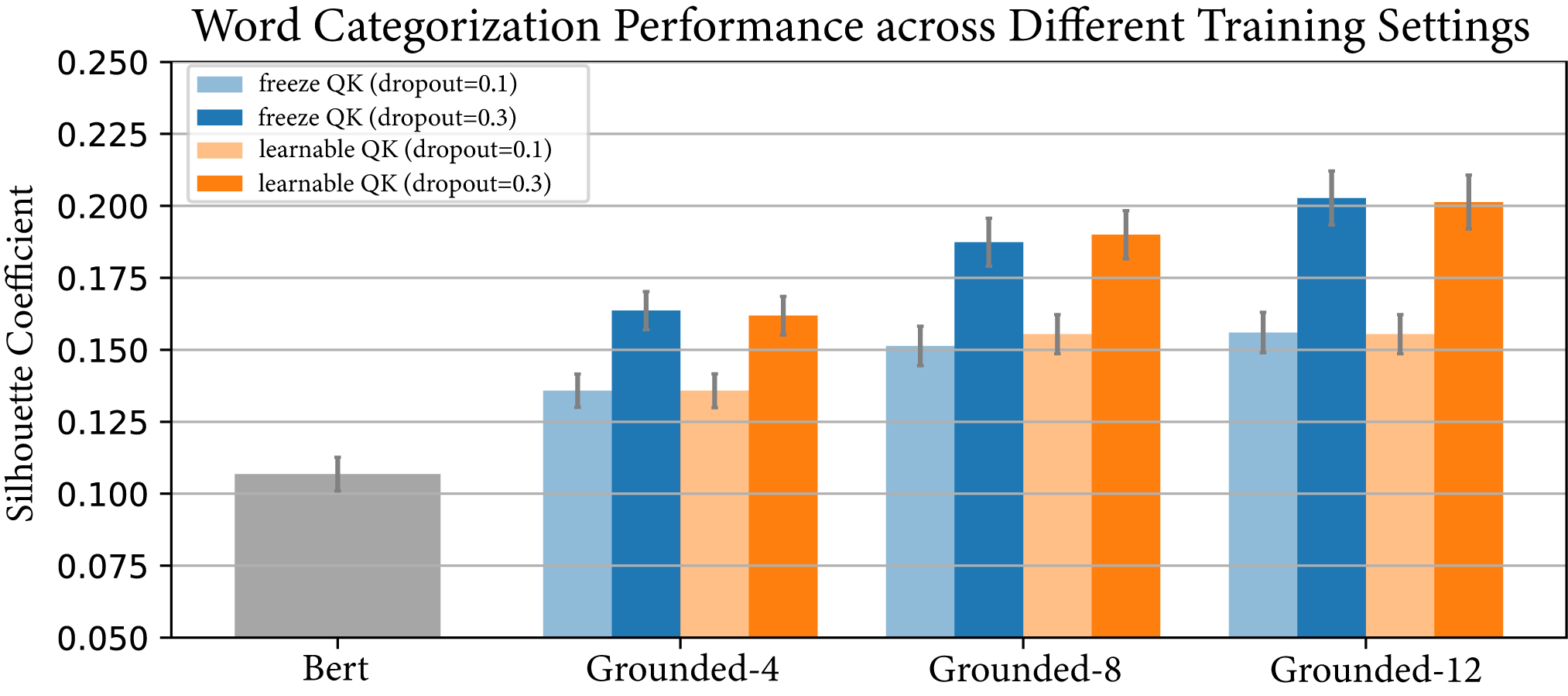}
  \caption{Word categorization performance compared across different training hyperparameters.}
  \label{fig:word_categorization_comparison}
\end{figure}

We also compare the word categorization performance after the visual grounding of natural language with different training settings (Fig.~\ref{fig:word_categorization_comparison}). The results suggest that earlier grounding (i.e. more learnable layers in Bert) tends to show better clustering performance on human-defined word categories. The models with frozen query and key transformations in Bert self-attention layers (blue bars) show similar categorization performance as the ones with learnable query and key weights for cross-modal training (orange bars). Models trained with a larger dropout rate ($0.3$; as shown in opaque bars) have significantly higher performance on word categorization than its counterpart with a smaller dropout rate ($0.1$; as shown in transparent bars).

\subsubsection{Word categorization performance is uncorrelated with its occurrence rate}

\begin{figure}[H]
  \centering
  \includegraphics[width=0.7\textwidth]{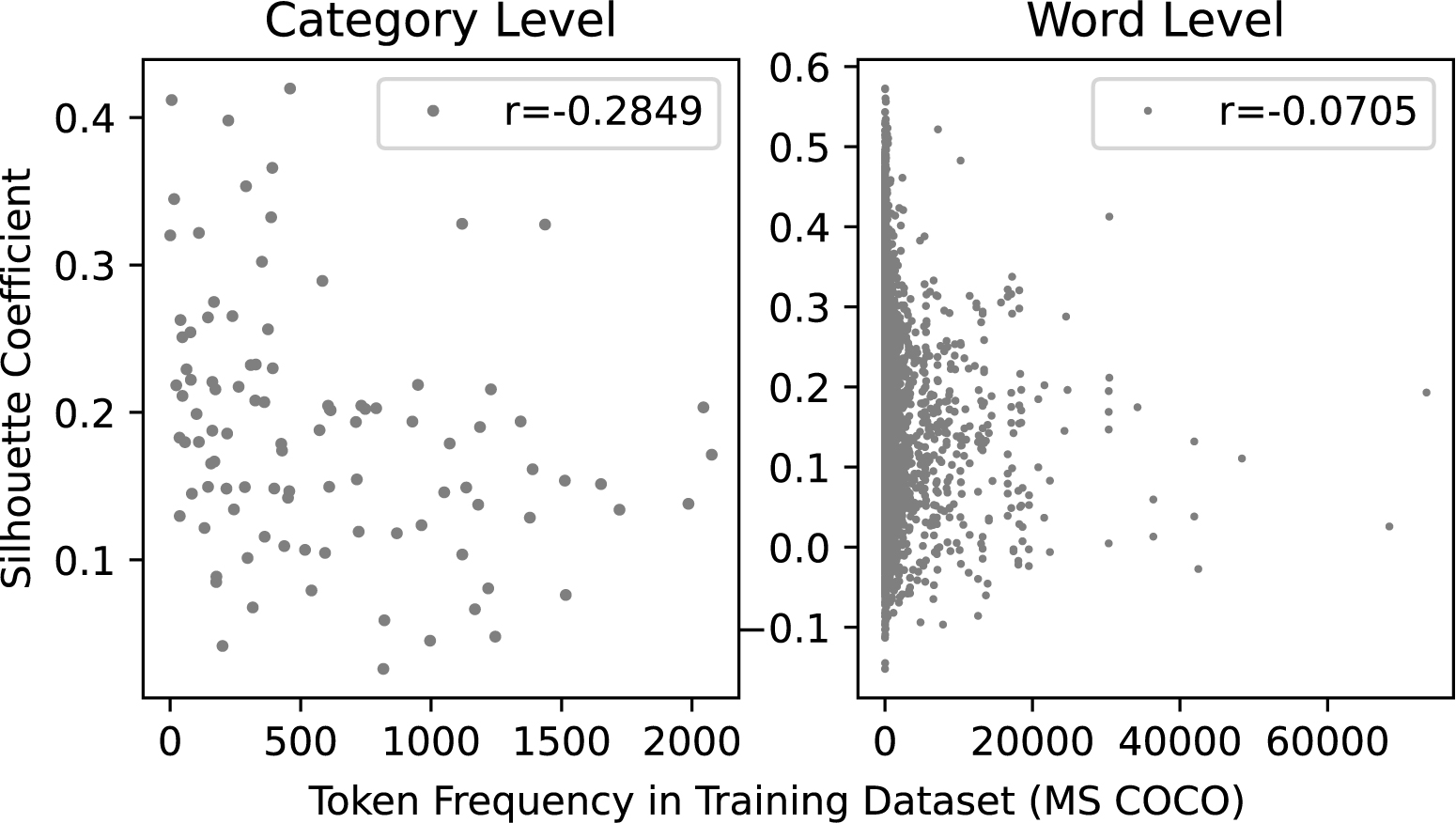}
  \caption{Word categorization performance is uncorrelated with its occurrence rate during training.}
  \label{fig:word_clustering_vs_frequency}
\end{figure}

In order to validate whether a better clustering performance on word $w_i$ results from a higher sampling rate in the training dataset, we further calculate the correlation between the Silhouette coefficient $s(i)$ and the training occurrence rate of word $w_i$ for all words in the SemCat dataset. The result (Fig.~\ref{fig:word_clustering_vs_frequency}) rejects this hypothesis by showing a weak correlation value between these two terms for both category-level ($r=-0.28$) and word-level ($r=-0.07$) analysis.

\newpage
\subsubsection{More examples on representations of word subcategories}

\begin{figure}[H]
  \centering
  \includegraphics[width=\textwidth]{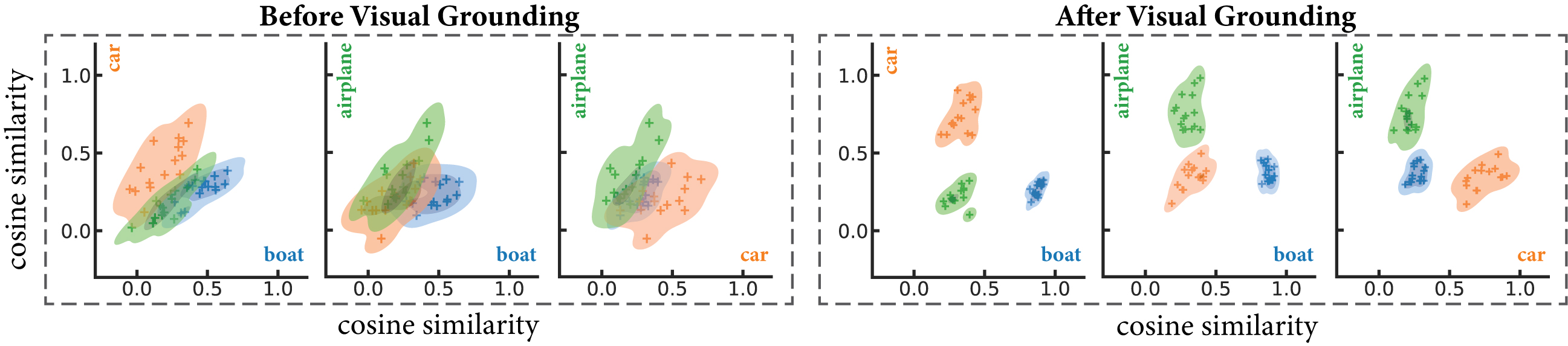}
  \caption{The distribution of representational similarity on \textbf{vehicle} words.}
  \label{fig:subclass_vehicle_ground}
\end{figure}

\begin{figure}[H]
  \centering
  \includegraphics[width=\textwidth]{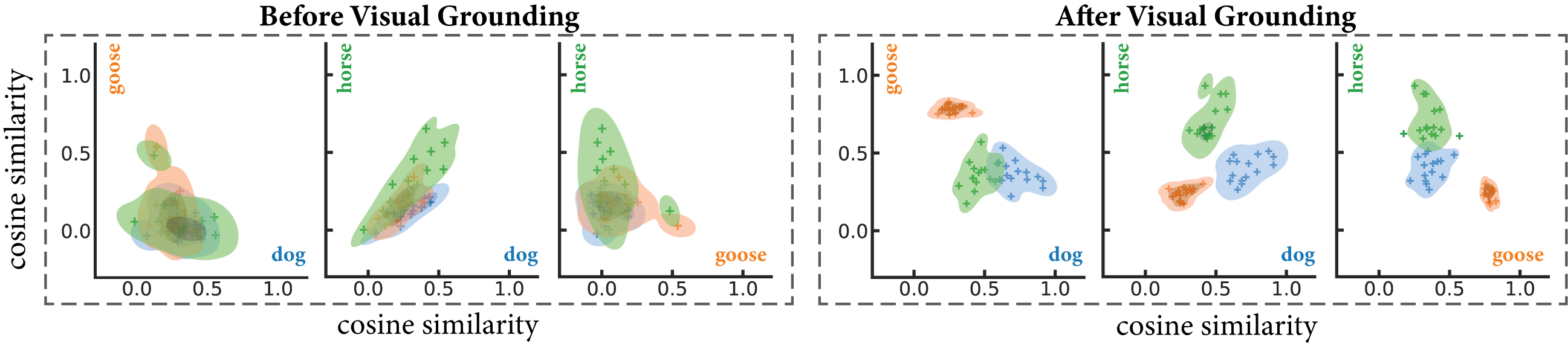}
  \caption{The distribution of representational similarity on \textbf{animal} words.}
  \label{fig:subclass_animal_ground}
\end{figure}

\begin{figure}[H]
  \centering
  \includegraphics[width=\textwidth]{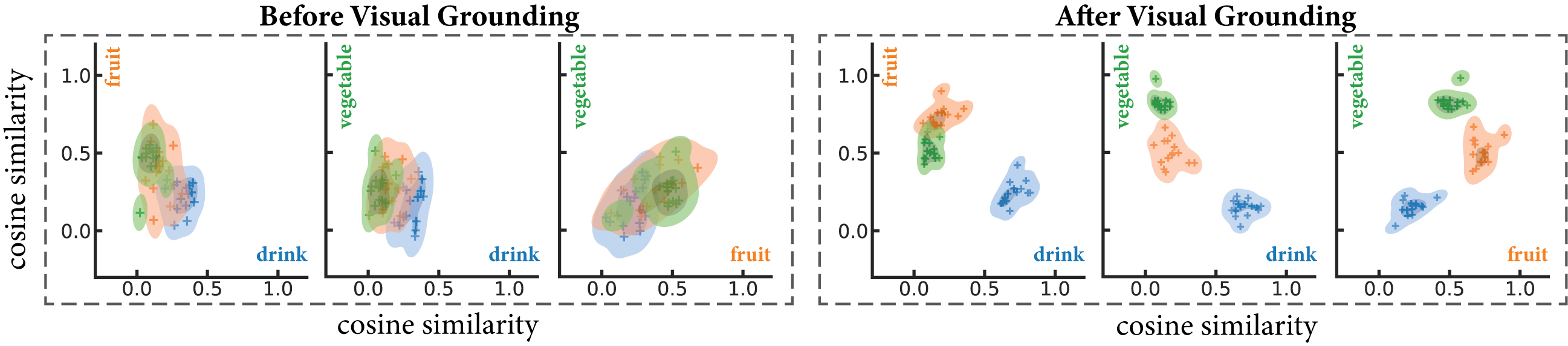}
  \caption{The distribution of representational similarity on \textbf{food} words.}
  \label{fig:subclass_food_ground}
\end{figure}

\begin{figure}[H]
  \centering
  \includegraphics[width=\textwidth]{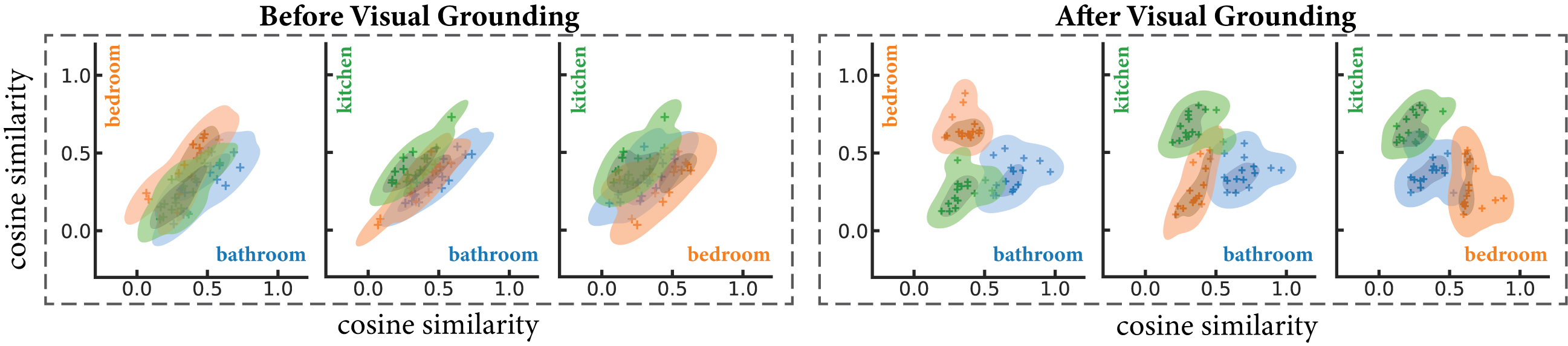}
  \caption{The distribution of representational similarity on \textbf{room} words.}
  \label{fig:subclass_room_ground}
\end{figure}

The following results are examples of \textbf{vehicle} (Table~\ref{table:subclass_vehicle}, Fig~\ref{fig:subclass_vehicle_ground}), \textbf{animal} (Table~\ref{table:subclass_animal}, Fig~\ref{fig:subclass_animal_ground}), \textbf{food} (Table~\ref{table:subclass_food}, Fig~\ref{fig:subclass_food_ground}), and \textbf{room} (Table~\ref{table:subclass_room}, Fig~\ref{fig:subclass_room_ground}) subcategories. For each subcategory, we first pick up three query words as the prototype for each subcategory (e.g. \texttt{boat}, \texttt{car}, \texttt{airplane} for \textbf{vehicle}). Then for each language model, we use the cosine similarity to sort out the top-$15$ closest words to each of these query words, as shown in the columns of these tables. We observe that the top similar words are well-aligned with the subcategory defined by the query word only after visual grounding. For example, all words in the column under "Grounded"/"Relational Grounded" for the \texttt{boat} query belong to water transportation, but words \emph{skeleton, nation, men, bike} found by Bert model are not water transportation. We further visualize the representation of the words in each subcategory from the "Grounded" model, by first calculating its cosine similarity to each of the query word (resulting in a three-dimensional vector), and then projecting these three-dimensional vectors into the 2D plane spanned by a pair of query word. The visualization results shown in the following figures further demonstrate that the subcategories are separable only after visual grounding.

\begin{table}[H]
  \small
  \caption{Top-15 words for \textbf{vehicle} subcategories}
  \label{table:subclass_vehicle}
  \centering
  \begin{tabular}{m{1.5cm} | m{3.2cm} | m{3.2cm} | m{3.2cm}}
    \toprule
    query       & Bert        & Grounded      & Relational Grounded \\
    \midrule
    boat   & \texttt{canoe, sailboat, submarine, skeleton, nation, sailing, men, bike, ballast, boating, raft, motorboat, paddle, yacht, kayak} & \texttt{tugboat, yacht, riverboat, gunboat, canoe, boating, barge,dinghy, raft, sailboat, ship, steamboat, steamer, trawler, watercraft} & \texttt{riverboat, gunboat, canoe, sailboat, dinghy, tugboat, yacht, motorboat, ship, steamer, barge, steamboat, submarine, ferry, steamship} \\
    \midrule
    car    & \texttt{vehicle, jeep, sedan, truck, jaguar, auto, driver, motorcycle, chassis, motor, bike, boat, horse, speeding, automobile} & \texttt{sedan, suv, vehicle, automobile, limo, jeep, limousine, taxi, drive, traffic, auto, pickup, van, roadster, truck} & \texttt{sedan, limo, automobile, suv, jeep, van, limousine, taxi, truck, buggy, vehicle, cart, motorcycle, auto, convertible} \\
    \midrule
    airplane    & \texttt{plane, aircraft, propeller, automobile, airport, bird, flight, parrot, turbulence, fly, kite, butterfly, rocket, motorcycle, takeoff} & \texttt{plane, jet, flight, aircraft, takeoff, airport, corsair, pilot, hangar, undercarriage, missile, propeller, nuclear, nautical, flyby} & \texttt{plane, jet, aircraft, corsair, flight, balloon, missile, takeoff, automobile, cockpit, hangar, avian, rocket, freighter, nuclear} \\
    \bottomrule
  \end{tabular}
\end{table}

\newpage

\begin{table}[H]
  \small
  \caption{Top-15 words for \textbf{animal} subcategories}
  \label{table:subclass_animal}
  \centering
  \begin{tabular}{m{1.5cm} | m{3.2cm} | m{3.2cm} | m{3.2cm}}
    \toprule
    query       & Bert        & Grounded      & Relational Grounded \\
    \midrule
    dog   & \texttt{pig, animal, bike, horse, mule, donkey, squirrel, bicycle, goat, monkey, motorcycle, moose, cat, gorilla, mouse} & \texttt{puppy, doge, terrier, canine, pug, hound, beagle, bulldog, dogwood, pup, mutt, spaniel, chihuahua, retriever, shepherd} & \texttt{puppy, doge, terrier, pug, canine, bulldog, beagle, hound, mutt, greyhound, bobcat, spaniel, hag, tomcat, donkey} \\
    \midrule
    goose    & \texttt{scare, geese, cow, calf, neighbor, puddle, flu, battleship, hog, displeasure, plank, herring, stir, scrambled, sock} & \texttt{geese, eagle, rooster, gull, pigeon, owl, duck, crow, parrot, partridge, falcon, harrier, sparrow, vulture, warbler} & \texttt{geese, eagle, pigeon, owl, parrot, duck, sparrow, rooster, falcon, crow, seagull, gull, warbler, partridge, harrier} \\
    \midrule
    horse    & \texttt{stallion, mule, dog, bike, trainer, boat, mare, car, animal, men, motorcycle, human, mountain, athlete, chestnut} & \texttt{stallion, mule, mare, donkey, seahorse, ox, steer, bull, unicorn, chestnut, foal, camel, oxbow, antelope, cow} & \texttt{stallion, mule, mare, seahorse, donkey, camel, bull, cow, cattle, ox, bison, deer, lassie, dog, animal} \\
    \bottomrule
  \end{tabular}
\end{table}

\newpage

\begin{table}[H]
  \small
  \caption{Top-15 words for \textbf{food} subcategories}
  \label{table:subclass_food}
  \centering
  \begin{tabular}{m{1.5cm} | m{3.2cm} | m{3.2cm} | m{3.2cm}}
    \toprule
    query       & Bert        & Grounded      & Relational Grounded \\
    \midrule
    drink   & \texttt{eat, feed, swallow, spend, study, spill, breathe, treat, bite, cough, drop, give, relax, wash, bleed} & \texttt{beverage, soda, cola, coke, juice, rum, bottle, champagne, drunk, flask, blender, mug, cup, blend, coffee} & \texttt{beverage, soda, cola, coke, juice, rum, flask, bottle, lemonade, coffee, cup, blend, mug, blender, brew} \\
    \midrule
    fruit    & \texttt{flower, foliage, citrus, flowers, plant, orchid, shrub, poisonous, inflorescence, eggs, omnivorous, seedling, nut, pineapple, snail} & \texttt{citrus, grape, strawberry, pear, pineapple, lemon, grapefruit, peach, mango, nuts, seeds, tomato,beets, tangerine, apple} & \texttt{citrus, strawberry, pineapple, grape, grapefruit, lemon, mango, peach, pear, apple, ripe, nuts, seeds, cherry, cranberry} \\
    \midrule
    vegetable    & \texttt{potato, beans, vegetables, cheese, mustard, beef, boiled, chicken, tomato, milk, corn, pig, bread, grape, grains} & \texttt{vegetables, salad, greens, botany, plantain,weeds, crops, algae, herb, legumes, perennial, lettuce, sprouts, vegetation, sprout} & \texttt{vegetables, botany, greens, legumes, salad, herb, plantain, herbs, lettuce, celery, crops, pomegranate, algae, asparagus, carrot} \\
    \bottomrule
  \end{tabular}
\end{table}

\newpage

\begin{table}[H]
  \small
  \caption{Top-15 words for \textbf{room} subcategories}
  \label{table:subclass_room}
  \centering
  \begin{tabular}{m{1.5cm} | m{3.2cm} | m{3.2cm} | m{3.2cm}}
    \toprule
    query       & Bert        & Grounded      & Relational Grounded \\
    \midrule
    bathroom   & \texttt{restroom, bedroom, bath, kitchen, toilet, laundry, refrigerator, couch, bathtub, dresser, shower, hallway, mirror, towel, backyard} & \texttt{restroom, bath, shower, bathtub, toilet, sink, vanity, wash, tub, mirror, towel, soap, hygiene, hallway, shave} & \texttt{restroom, kitchen, cafeteria, bedroom, bath, room, shower, hospital, office, gym, classroom, pantry, hotel, gymnasium, motel} \\
    \midrule
    bedroom    & \texttt{bathroom, bed, room, dresser, downstairs, apartment, kitchen, condo, couch, upstairs, backyard, mattress, hallway, bath, attic} & \texttt{bed, mattress, pillow, room, closet, condominium, crib, dresser, apartment, motel, upstairs, cot, dorm, blanket, robe} & \texttt{room, closet, dorm, hotel, kitchen, apartment, motel, bathroom, household, dormitory, office, hostel, classroom, cafeteria, house} \\
    \midrule
    kitchen    & \texttt{refrigerator, bathroom, couch, fireplace, backyard, laundry, barn, furniture, sofa, basement, toilet, bedroom, cupboard, stairs, driveway} & \texttt{pantry, counter, cupboard, household, galley, stove, cook, microwave, oven, refrigerator, freezer, kettle, furnace, washer, chef} & \texttt{pantry, cafeteria, household, bathroom, bedroom, restroom, room, office, showroom, classroom, restaurant, garage, parlor, gym, dugout} \\
    \bottomrule
  \end{tabular}
\end{table}

\newpage
\subsection{Extended results on vision based concept composition}

\begin{figure}[H]
  \centering
  \includegraphics[width=\textwidth]{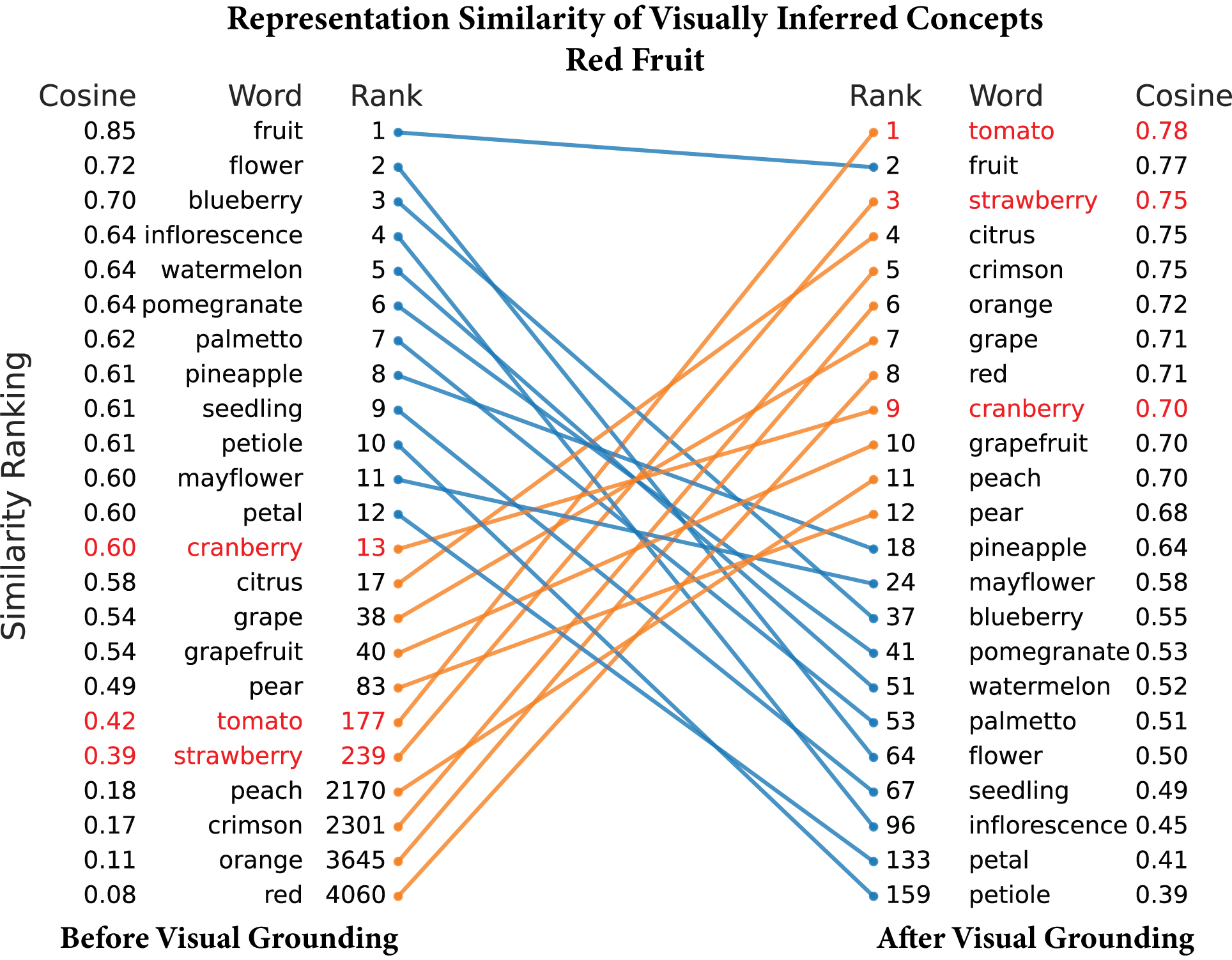}
  \caption{Concept composition based on visual knowledge (\textbf{red fruit}). }
  \label{fig:visual_inference_red_fruit}
\end{figure}

For evaluating the vision-based concept composition described in main text Section 4.4, we also visualize the ranking change of words that are most similar to a query phrase before and after visual grounding with a slope chart. In addition to the \texttt{striped horse} example shown in the main text, we also plot the slope chart for example query \texttt{red fruit} (Fig.~\ref{fig:visual_inference_red_fruit}). The result suggests both models found most similar words belonging to fruit/plant, but after grounding the color information learned from the visual stream enhance the semantic representation of words like \emph{tomato, strawberry, cranberry} to be closer to \texttt{red fruit}.

\newpage
\subsection{Supplementary method for multimodal image search}

To implement multimodal image search, we first add two additional heads ($F_V$ and $F_L$) to the image key and langauge query (after being concatenated across all attention heads along the feature dimension) from the cross-modal attention module in Section 3.2.

\begin{equation}
    \label{eq:cross_modal_search_head_V}
    \vb*{Q}_I = F_V(\text{Key}_V) = \frac{1}{HW}\sum_{i,j} \Big( \big(\text{ReLU}(\text{Key}_V [i,j,:] \vb*{W}_V^1+\vb*{b}_V^1) \big) \vb*{W}_V^2+\vb*{b}_V^2 \Big),
\end{equation}

\begin{equation}
    \label{eq:cross_modal_search_head_L}
    \vb*{Q}_W = F_L(\text{Query}_L) = \frac{1}{K}\sum_{k} \Big( \big(\text{ReLU}(\text{Query}_L [k,:] \vb*{W}_L^1+\vb*{b}_L^1) \big) \vb*{W}_L^2+\vb*{b}_L^2 \Big),
\end{equation}

where $\vb*{W}$s and $\vb*{b}$s are the weights and biases of the linear transformations in these two head functions, with size $d \times d$ and $d \times 1$ ($d=768$). $H$ and $W$ are the height and width of the image feature output ($H=W=14$). $K$ is the number of words in an image caption, which varies for different language inputs. $F_V$ and $F_L$ are applied to visual and textual representations ($\text{Key}_V$ or $\text{Query}_L$) in the joint space respectively. After average pooling the outputs, we get a single vector representation for either an image (denoted as $\vb*{Q}_I \in \mathbb{R}^d$) or a text (denoted as $\vb*{Q}_W \in \mathbb{R}^d$). 

These two representations are L2 normalized (Eq.~\ref{eq:L2_norm_query}) and aligned into a shared space, by freezing the pretrained two-stream model and only finetuning the transformation heads $F_V$ and $F_L$ with contrastive loss to match paired images and texts in terms of their cosine similarity using the MS COCO dataset. 

\begin{equation}
    \label{eq:L2_norm_query}
    \vb*{Q}_I \leftarrow \frac{\vb*{Q}_I}{\|\vb*{Q}_I\|_2}, \quad \vb*{Q}_W \leftarrow \frac{\vb*{Q}_W}{\|\vb*{Q}_W\|_2},
\end{equation}

\begin{equation}
    \label{eq:cross_modal_search_query}
    \vb*{Q}_{\text{search}} = (1-\alpha) \vb*{Q}_I + \alpha \vb*{Q}_W.
\end{equation}

Then to construct a multimodal search query $\vb*{Q}_{\text{search}}$, we use a linear combination of a pair of image query $\vb*{Q}_I$ and text query $\vb*{Q}_W$ (since they are now in the same representational space) weighted by a scalar $\alpha$ (Eq.~\ref{eq:cross_modal_search_query}). When $\alpha=0$, only the image is used as a query for search and retrieval. When $\alpha=1$, only the word is used as the search query. And when $\alpha=0.5$, the image and text inputs contribute equally to the search query. The weight $\alpha$ can vary continuously in the range from $0$ to $1$. If the visual and text-informed semantics share a common semantic space, then varying alpha is expected to result in the retrieved images to vary their contents according to progressive and intuitive transition from the image to the text. 

\newpage
\subsection{Evaluate the grounded language model on GLUE benchmark}

Similar to the evaluation in \citep{devlin2019bert}, we further test the language model before and after visual grounding on the General Language Understanding Evaluation (GLUE) benchmark \citep{wang2018glue}. For this purpose, the Bert encoder in the language models is fixed, while only a pooling layer (shared across all tasks in GLUE) and a linear classification layer (specific for each task in GLUE) are trainable. We use the training and evaluation codes from the \texttt{jiant} \footnote{Jiant package: \url{https://github.com/nyu-mll/jiant/}} package. The testing results are submitted to and evaluated by the GLUE benchmark website \footnote{\url{https://gluebenchmark.com/}}.

\begin{table}[H]
  \small
  \caption{Model performance on GLUE benchmark.}
  \label{table:GLUE_performance}
  \centering
  \begin{tabular}{ m{2cm} m{0.7cm} m{0.8cm} m{0.8cm} m{0.9cm} m{0.7cm} m{1.3cm} m{0.8cm} m{0.8cm} m{1cm} }
    \toprule
    Model           & CoLA    & SST-2   & MRPC    & STS-B   & QQP     & MNLI-(m/mm)     & QNLI    & RTE     & Average   \\ [10pt]
    \midrule
    Bert            & 40.7    & 92.6    & 87.8    & 81.8    & 71      & 83.3/82         & 89.4    & 73.8    & 78.04     \\ 
    \midrule
    Grounded-4      & 37.1    & 91.4    & 86      & 83.3    & 70.8    & 82.7/81.5       & 88.9    & 73      & 77.19     \\ 
    \midrule
    Grounded-8      & 38.6    & 91.5    & 86.3    & 83.4    & 70.8    & 82/81           & 87.9    & 72.2    & 77.08     \\ 
    \midrule
    Grounded-12     & 37.2    & 92.6    & 86.5    & 82.3    & 70.5    & 82.3/81.7       & 89.2    & 72      & 77.10     \\ 
    \midrule
    Relational-2       & 38      & 92.8    & 84.6    & 81.8    & 70.4    & 83.1/81.8       & 89.2    & 71.7    & 77.04   \\
    \bottomrule
  \end{tabular}
\end{table}

Table~\ref{table:GLUE_performance} summarizes the results. \texttt{Grounded-$k$} models have $k$ learnable layers in Bert for visual grounding of natural language with MS COCO dataset. \texttt{Relational-2} model is trained from \texttt{Grounded-$8$} by finetuning $2$ Bert layers for visual grounding of object relations.

The results suggest that in general the language model has a slightly decreased performance on GLUE after visual grounding, although the models tested here all share the same architecture as Bert. Allowing more learnable parameters during visual grounding tends to result in worse performance for natural language understanding. This is unsurprising since the grounding process is based on matching short captions (MS COCO dataset) or phrases (Visual Genome dataset) to visual contents, which may not require extensive capacity for textual processing. Future study is needed to reconcile the trade-off performance between visual grounding and natural language understanding.

\section{Code}
We release the code for training and testing the proposed models at \url{https://github.com/yizhen-zhang/VG-Bert}.

\newpage
\bibliographystyle{unsrtnat}
\small
\bibliography{supplement.bib}